\documentclass[10pt,journal,compsoc]{IEEEtran}
\pdfoutput=1
\usepackage{graphicx}
\usepackage{amsmath}
\usepackage{amssymb}
\usepackage{booktabs}

\usepackage{algorithm}
\usepackage{algorithmic}
\usepackage{multirow}
\usepackage[table,xcdraw]{xcolor}
\usepackage{amssymb}
\usepackage{bbding}
\usepackage{autobreak}

\usepackage[pagebackref,breaklinks,colorlinks]{hyperref}

\usepackage[capitalize]{cleveref}

\ifCLASSOPTIONcompsoc
  \usepackage[nocompress]{cite}
\else
  \usepackage{cite}
\fi

\ifCLASSINFOpdf
\else
\fi
\begin{document}
%
\title{Are Dense Labels Always Necessary for \\ 3D Object Detection from Point Cloud?}
%
%
%
%

\author{Chenqiang~Gao,  
        Chuandong~Liu, 
        Jun Shu, 
        Fangcen Liu,
        Jiang Liu,
        Luyu Yang,
        Xinbo Gao,
        and Deyu Meng

\IEEEcompsocitemizethanks{
\IEEEcompsocthanksitem C. Gao is with the School of Intelligent Systems Engineering, the Shenzhen Campus of Sun Yatsen University, Sun Yat-sen University, Shenzhen, Guangdong 518107, China. E-mail: gaochq6@mail.sysu.edu.cn. 
\IEEEcompsocthanksitem C. Liu is with the School of Computer Science, Wuhan University, Wuhan 430072, China. E-mail: chuandong.liu@whu.edu.cn.
\IEEEcompsocthanksitem J. Shu and D. Meng are with School of Mathematics and Statistics and Ministry of Education Key Lab of Intelligent Networks and Network Security, Xi’an Jiaotong University, Shaanxi 710049, China.
E-mail: xjtushujun@gmail.com, dymeng@mail.xjtu.edu.cn.
\IEEEcompsocthanksitem F. Liu and X. Gao are with School of Communication and Information Engineering, Chongqing University of Posts and Telecommunications, Chongqing 400065, China. E-mail: gaoxb@cqupt.edu.cn, liufc67@gmail.com. 
\IEEEcompsocthanksitem J. Liu is with Meta, Menlo Park 94025, U.S.
E-mail: jiangliu@meta.com.
\IEEEcompsocthanksitem L. Yang is with University of Maryland, College Park 20742, U.S. \protect\\
E-mail: loyo@umiacs.umd.edu.
}
}

%
%

\markboth{IEEE Transactions on Pattern Analysis and Machine Intelligence}
{IEEE Transactions on Pattern Analysis and Machine Intelligence}
\IEEEtitleabstractindextext{
\begin{abstract}
     Current state-of-the-art (SOTA) 3D object detection methods often require a large amount of 3D bounding box annotations for training. However, collecting such large-scale densely-supervised datasets is notoriously costly. To reduce the cumbersome data annotation process, we propose a novel sparsely-annotated framework, in which we just annotate one 3D object per scene. Such a sparse annotation strategy could significantly reduce the heavy annotation burden, while inexact and incomplete sparse supervision may severely deteriorate the detection performance. To address this issue, we develop the SS3D++ method that alternatively improves 3D detector training and confident fully-annotated scene generation in a unified learning scheme. Using sparse annotations as seeds, we progressively generate confident fully-annotated scenes based on designing a missing-annotated instance mining module and reliable background mining module. Our proposed method produces competitive results when compared with SOTA weakly-supervised methods using the same or even more annotation costs. Besides, compared with SOTA fully-supervised methods, we achieve on-par or even better performance on the KITTI dataset with about 5× less annotation cost, and 90\% of their performance on the Waymo dataset with about 15× less annotation cost. The additional unlabeled training scenes could further boost the performance. The code will be available at \href{https://github.com/gaocq/SS3D2}{https://github.com/gaocq/SS3D2}.
\end{abstract}

}
\maketitle

\IEEEdisplaynontitleabstractindextext


%
\IEEEpeerreviewmaketitle

\IEEEraisesectionheading{\section{Introduction}\label{sec:introduction}}

%
%
%
%

\IEEEPARstart{A}{utonomous} driving, which aims to enable vehicles to perceive the surrounding environments intelligently, has attracted much attention recently \cite{mao20233d}. As one of the fundamental problems in autonomous driving, detecting 3D objects accurately from a point cloud is crucial. 
With the development of deep learning techniques, a number of approaches \cite{pointrcnn,voxelrcnn,sessd,ciassd} based on either voxel-wise or point-wise features have been proposed and achieved state-of-the-art (SOTA) performance on large-scale datasets \cite{kitti,waymo}. However, most of the proposed methods require precisely densely-annotated datasets. Unfortunately, collecting high-quality 3D annotations is often time-consuming and labor-intensive, requiring hundreds of hours to annotate just one hour of driving data~\cite{weaklysuper}. Therefore, it is necessary and significant to reduce the cumbersome annotation process.

\begin{figure}[t] 
	\centering
	\includegraphics[width=\linewidth]{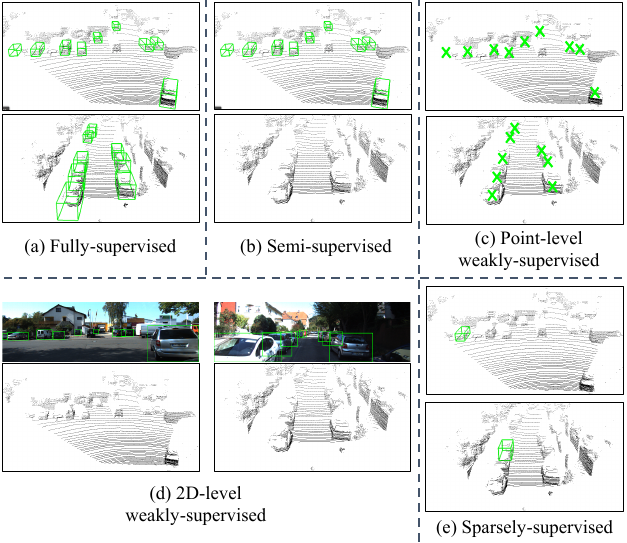}
    \put(-180.5,117){\fontsize{6pt}{6pt}\selectfont \cite{pointrcnn}}
    \put(-95.5,117){\fontsize{6pt}{6pt}\selectfont \cite{3dioumatch}}
    \put(-11.5,112){\fontsize{6pt}{6pt}\selectfont \cite{weaklysuper}}
    \put(-144,8.5){\fontsize{6pt}{6pt}\selectfont \cite{fgr}}
	\caption{
		Illustration of annotation setup for different supervision forms.
		The green boxes represent 3D or 2D box annotations and the green crosses represent point-level center annotations in the BEV point cloud map. In this paper, we explore a sparse annotation setup in which we just annotate one 3D object in each scene, as shown in Fig. (e).
	}
	\label{fig:diffierent} 
\end{figure}

\begin{figure}[t]
	\centering
	\includegraphics[width=\linewidth]{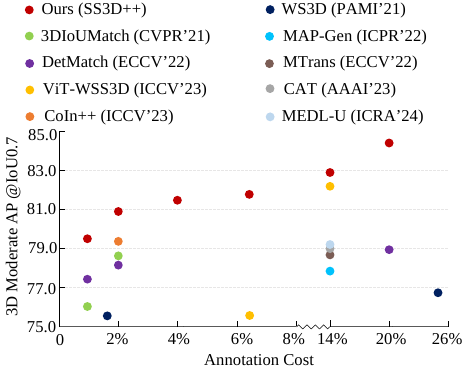}
    \put(-139,178){\fontsize{9pt}{9pt}\selectfont \cite{3dioumatch}}
    \put(-148,162){\fontsize{9pt}{9pt}\selectfont \cite{park2022detmatch}}
    \put(-140,148){\fontsize{9pt}{9pt}\selectfont \cite{wss3d}}
    \put(-155,133){\fontsize{9pt}{9pt}\selectfont \cite{xia2023coin}}
    \put(-34,191){\fontsize{9pt}{9pt}\selectfont \cite{weaklysuper}}
    \put(-22,178){\fontsize{9pt}{9pt}\selectfont \cite{liu2022mapgen}}
    \put(-30,162){\fontsize{9pt}{9pt}\selectfont \cite{mtrans}}
    \put(-39,148){\fontsize{9pt}{9pt}\selectfont \cite{cat}}
    \put(-25,133){\fontsize{9pt}{9pt}\selectfont \cite{medl}}
	\caption{
		Illustration of 3D moderate AP (Average Precision) \emph{vs.} annotation cost, tested on KITTI ~\cite{kitti} validation set and the AP is calculated with 40 recall positions at IoU 0.7 for the car to compare with the previous methods. Compared with SOTA semi-supervised detectors, our SS3D++ yields promising results with a far lower annotation demand. Besides, when providing less annotations than the remarkably weakly-supervised methods, our SS3D++ still shows profitable detection performance. The cost calculation is based on analysis in FGR \cite{fgr} and ViT-WSS3D \cite{wss3d}.
	}
	\label{fig:fig_comparison}  
\end{figure}
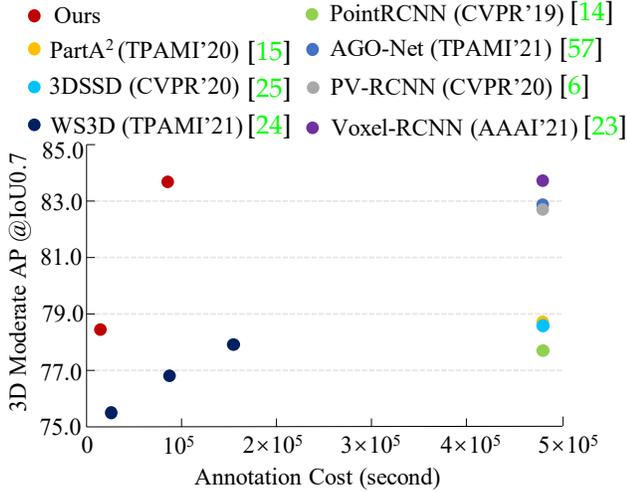

To mitigate this challenge, some weakly-supervised and semi-supervised approaches \cite{fgr,3dioumatch,vs3d} have been proposed, which attempt to learn 3D detector from relatively easily acquired annotations, e.g, leveraging a few labeled and much unlabeled data \cite{3dioumatch}, center-click point \cite{weaklysuper} or 2D image-level bounding box annotation~\cite{fgr}, as depicted in Fig. \ref{fig:diffierent}. Though these methods have substantially reduced the heavy annotation burden, there exist several deficiencies for practice. On the one hand, their performances are unsatisfied and would not meet the requirements for practical applications.
On the other hand, these low-cost annotations are hardly directly trained using fully-supervised detectors.
These specific detector designs for weakly-supervised forms relatively restrict the potential to benefit from up-to-date fully-supervised detectors.
Both of the above two aspects may prevent the applicability in real-world settings.

To promote the deployment of 3D detector systems, we propose a novel sparsely-supervised method that learns 3D detectors from sparse annotation in which we just annotate one 3D object per scene, as show in Fig. \ref{fig:diffierent}. 
Intuitively, such novel sparse annotation form facilitates learning information of unlabeled objects, since intra-scene information transfer is much easier than inter-scene information transfer. Comparatively, semi-supervised scenario \cite{3dioumatch} may fall into the suboptimal solution due to limited information transfer from labeled to large discrepancy unlabeled scenes, and existing weakly-supervised methods may suffer intractable location issues due to significant disparity between 2D-level \cite{fgr} or point-level \cite{weaklysuper} weak annotations and 3D-level annotations. Therefore, the proposed sparse annotation has the potential to achieve better performance.

However, such sparse annotation data provides inexact and incomplete supervision, naturally raising several new challenges.
(1) The missing-annotated instances and the region near those instances may be incorrectly marked as background, which can easily deteriorate the detector under the guidance of incorrect negative samples;
(2) To eliminate the negative effect of inexact supervision from positive samples, we are required to generate reliable pseudo-annotated instances (high precision), and find possibly many newly annotated instances (high recall);
(3) The sparsely-annotated object in each scene is randomly chosen, while the hardness of learning different objects has great differences. This possibly hinders the knowledge transfer from labeled to unlabeled objects, and impairs the performance.

To address these challenges, we propose an effective method for sparsely annotated 3D object detection, namely SS3D++.   
The main idea of SS3D++ is to progressively mine positive instances and backgrounds with high confidence using our reliable background mining and missing-annotated instance mining modules, and further use them to generate confident fully-annotated scenes by the GT sampling \cite{second} data augmentation strategy. 
Such a strategy guarantees that generated pseudo annotations of training scenes with high recall, providing sufficient knowledge.

Our sparsely-supervised framework has three appealing characteristics. First, we can achieve high performance or even surpassing SOTA fully-supervised methods, as presented in Tab. \ref{tab:tab1} and Tab. \ref{tab:tab1waymo} about experimental results of the KITTI~\cite{kitti} and Waymo datasets~\cite{waymo}. However, we only require $5\times$ less annotations than fully-supervised methods on KITTI dataset \cite{kitti}, as shown in Fig. \ref{fig:fig_comparison}. 
Besides, when our annotation cost is similar or even less than existing weakly-supervised approaches~\cite{fgr,weaklysuper,3dioumatch}, we can achieve higher results.
Specifically, under the similar number of 3D annotation objects, our annotations are relatively more easily acquired than existing weakly-supervised methods, since it is often easy to just annotate one 3D object per scene.

Second, our method is detector-agnostic, which can easily benefit from advanced fully-supervised 3D detectors. 
As illustrated in Tables \ref{tab:tab1} and \ref{tab:tab1waymo}, we instantiate our SS3D++ framework as these fully-supervised detectors without modifying the architectures, which consistently improves the performance approaching or even surpassing the fully-supervised results. 
This property is promising to apply our framework into the autonomous driving system, since it is simple and convenient to adapt to up-to-date 3D detectors and refresh the system to obtain better performance. However, existing weakly-supervised methods \cite{fgr,weaklysuper} elaborately designing specific architectures of detectors to address such imperfect information, relatively hardly benefit from advanced fully-supervised 3D detectors.

Third, our SS3D++ method is hopeful to learn from additional unlabeled training scenes. On the one hand, if we have access to additional unlabeled training scenes in an offline manner, as presented in Fig. \ref{fig:com_semi}, we can further boost the performance of a 3D detector by making use of these unlabeled data. This implies that our method is capable of generating confident pseudo annotations for easily obtained unlabeled scenes. On the other hand, if additional unlabeled training scenes arrive in an online manner, we can also continuously improve the performance of the 3D detector by making use of these streaming unlabeled data, as shown in Fig. \ref{fig:three_fig} (c). This property is potentially useful for autonomous driving system, which is required to perceive the changing environment and incrementally improve itself.

This work builds upon our conference paper~\cite{liu2022ss3d} which has attracted promising attention in the research community. Subsequent studies have further advanced sparse supervision through various approaches, including prototype-based object mining~\cite{zhu2025learning}, cross-modal prompts~\cite{zhao2025sp3d}, hard instance enhancement~\cite{HINTED}, and contrastive instance feature mining~\cite{xia2023coin}. In contrast, the work in this paper significantly extends the original conference paper in various aspects.
Firstly, we introduce the hardness of learning different objects to eliminate the influence of randomly selecting a sparse-annotated object in a scene in Sec. \ref{criteria}.
Secondly, we rigorously formulate the missing positive instances mining as a multi-criteria sample selection process, and use sample selection curriculum setups to adaptively determine high-confidence positive instances involved in the training pool in Sec. \ref{foreground}. Thirdly, we improve the algorithm by leveraging the mutual benefit between 3D detector training and confident fully-annotated scene generation in Sec. \ref{scene}. Fourthly, extensive ablation studies (see Sec. \ref{ablation}) and experiments on large-scale Waymo dataset (see Tab. \ref{tab:tab1waymo} and {Tab. \ref{tab:waymo_result_semi}}) are conducted to validate the effectiveness and universality. Fifthly, we make an in-depth comparison with other weakly-supervised methods in Sec. \ref{weakly}. Finally, we show that the proposed method has the potential for utilizing additional unlabeled scenes to ameliorate the model in Sec. \ref{unlabel}.

\section{Related Work}
\textbf{Fully-supervised 3D object detection.}
The existing 3D detection methods can be broadly categorized into two types: voxel-based methods \cite{second,sessd,ciassd,pointpillars,sassd,Wang_2023_CVPR,zhang2023glenet,lan2022arm3d} and point-based methods \cite{f-pointnet,std,3dssd,vote,pointgnn,pcrgnn}.
Thanks to the seminal PointNet series~\cite{pointnet,pointnet++}, point-based methods directly take the raw irregular points as input to extract local and global features.
PointRCNN~\cite{pointrcnn} directly generated 3D proposals from raw points in a bottom-up manner.
Part-A$^2$~\cite{parta2} further improved the PointRCNN by exploring the rich information in intra-object part locations.
For voxel-based methods, voxelization is a common measure for irregular point clouds to apply traditional 2D or 3D convolution.
PV-RCNN~\cite{pvrcnn} extended the SECOND~\cite{second} by leveraging a keypoints branch to learn discriminative features.
PV-RCNN++~\cite{pvrcnn++} proposed the VectorPool aggregation and sectorized keypoints sampling strategy to further enhance the PV-RCNN.
Voxel-RCNN~\cite{voxelrcnn} achieved a careful balance between performance and efficiency based on the voxel representations.
CenterPoint~\cite{centerpoint} firstly used a keypoint detector to detect centers of objects and then regressed to other attributes.
MPPNet \cite{chen2022mppnet} leveraged the online tracker to link 3D objects in the temporal dimension.
Meanwhile, some methods \cite{qi2021offboard,ma2023detzero} explored the rich temporal information among frames to facilitate the 3D object detection.
SST \cite{sst} and DSVT \cite{dsvt} utilized transformers as replacements for sparse CNNs.

Prior works have made significant progress and achieved impressive performance, while these results deeply depend on the large-scale manually precisely densely-annotated 3D data, which are time-consuming and labor-intensive. 
We make an attempt towards weakly supervised 3D object detection considering the sparse annotation strategy which just annotates one object per scene. This enables a much faster and easier data labeling process compared to notoriously densely full supervision, and we can achieve comparable or better performance.
Moreover, our method is detector-agnostic, which can easily benefit from advanced fully-supervised voxel-based or point-based detectors. We will instantiate the SS3D++ as above fully-supervised 3D detectors to demonstrate the effectiveness of our method in the experimental section.

\noindent\textbf{Weakly/Semi-supervised 3D object detection.}
To reduce annotations of 3D objects, the weakly-supervised learning strategy is adopted in WS3D \cite{weaklysuper}, which is achieved by a two-stage architecture based on the click-annotated scheme. WS3D~\cite{weaklysuper} generated cylindrical object proposals by click-annotated scenes in stage-1 and refined the proposals to get cuboids using slight well-labeled instances in stage-2. 
ViT-WSS3D \cite{wss3d} studied weakly semi-supervised 3D detection with massive click-annotated data and a small number of fully labeled 3D objects. In addition to the weakly-supervised methods based on the point-level annotations, there are also methods \cite{fgr,liu2022mapgen,cat} that utilize 2D bounding boxes for alleviating the cumbersome 3D annotation process. MTrans \cite{mtrans} proposed a cross-modal Transformer to complement point cloud with dense visual images. Following MTrans, CAT \cite{cat} introduced a global encoder to interpret contextual information, thereby removing the visual content. Further, MEDL-U \cite{medl} introduced an evidential deep learning based uncertainty estimation framework to address the ambiguities in pseudo labels. Though point-level and 2D-level weakly supervised methods have subtantially reduced the heavy data annotation, their dissatisfied performances and elaborately designed structure may prevent the applicability.
Besides, the supervision provided by the weakly-supervised point-level or 2D-level annotation is too weak, so that a certain amount of full annotations have to be provided additionally. 

Meanwhile, some semi-supervised methods~\cite{sess,park2022detmatch,yin2022semi,mt} leveraged the mutual teacher-student framework to transfer information from labeled scenes to unlabeled scenes.
3DIoUMatch~\cite{3dioumatch} conducted a pioneering work for outdoor scenes, which estimate 3D IoU as a localization metric and set a self-adjusted threshold to filter pseudo labels.
HSSDA \cite{liu2023hierarchical} leveraged the hierarchical supervision and shuffle data augmentation to facilitate the semi-supervised learning. A-teacher \cite{ateacher} designed an online asymmetric framework with attention-based refinement to improve the quality of pseudo labels. PPTM \cite{pptm} generated strong pseudo labels with cross-mixes pillars data augmentation. 
Though these methods have substantially reduced the heavy data annotation burden, they are relatively hard to benefit from advanced 3D detectors. Besides, the semi-supervised methods need to annotated all the 3D objects within a scene, in which annotators have to switch viewpoints or zoom in and out throughout a 3D scene carefully. 
All of these prevent potential practical applicability of 3D object detection. In this paper, we attempt to explore a new weakly-supervised form, sparse supervision, and facilitate 3D object detection close to practical applications.
The proposed sparse annotation, which requires labeling only a subset of objects, significantly reduces the number of point clouds that need attention. This reduction not only lessens the burden on annotators but also minimizes the potential for errors during the annotation process, thereby greatly lowering the costs associated with manual work.

\noindent\textbf{Sparsely-supervised 2D object detection.}
Due to that a part of instances are missing-annotated, weight updated of the network may be misguided significantly when gradients back-propagated. 
To address this issue, existing 2D detection methods employed re-weight or re-calibrates strategy on the loss of RoIs (regions of interest) to eliminate the effect of unlabeled instances. 
Soft sampling \cite{softsamp} utilized overlaps between RoIs and annotated instances to re-weight the loss.
Background recalibration loss\cite{solving} based on focal loss\cite{focal} regarded the unlabeled instances as hard-negative samples and re-calibrates their losses, which is only applicable to single-stage detectors. 
Especially, part-aware sampling\cite{partsampling} ignored the classification loss for part categories by using human intuition for the hierarchical relation between labeled and unlabeled instances. 
Co-mining \cite{comin} proposed a co-generation module to convert the unlabeled instances as positive supervisions.
However, these sparsely annotated object detection methods are all for 2D image objects.
Due to the modal difference between 2D images and 3D point cloud, these methods can not be directly applied to our 3D object detection task.
For example, in KITTI \cite{kitti}, 3D objects are naturally separated, which means the overlaps among objects are zero and the hierarchical relation between objects does not exist. 
Compare with the re-weight and re-calibrates methods, in this paper, we propose a novel method for sparsely annotated 3D object detection which leverages a simple but effective background mining module and a missing-annotated instance mining module to mine confident positive instances and backgrounds, which is key for training detectors with high performance.

\noindent\textbf{Curriculum learning for object detection.}
Due to that a part of instances are missing-annotated, weight updated of the network may be misguided significantly when gradients back-propagated. 
To address this issue, existing 2D detection methods employed re-weight or re-calibrates strategy on the loss of RoIs (regions of interest) to eliminate the effect of unlabeled instances. 
Soft sampling \cite{softsamp} utilized overlaps between RoIs and annotated instances to re-weight the loss.
Background recalibration loss\cite{solving} based on focal loss\cite{focal} regarded the unlabeled instances as hard-negative samples and re-calibrates their losses, which is only applicable to single-stage detectors. 
Especially, part-aware sampling\cite{partsampling} ignored the classification loss for part categories by using human intuition for the hierarchical relation between labeled and unlabeled instances. 
Co-mining \cite{comin} proposed a co-generation module to convert the unlabeled instances as positive supervisions.
However, these sparsely annotated object detection methods are all for 2D image objects.
Due to the modal difference between 2D images and 3D point cloud, these methods can not be directly applied to our 3D object detection task.
For example, in KITTI \cite{kitti}, 3D objects are naturally separated, which means the overlaps among objects are zero and the hierarchical relation between objects does not exist. 
Compare with the re-weight and re-calibrates methods, in this paper, we propose a novel method for sparsely annotated 3D object detection which leverages a simple but effective background mining module and a missing-annotated instance mining module to mine confident positive instances and backgrounds, which is key for training detectors with high performance.

\noindent\textbf{Curriculum learning for object detection.}
Curriculum learning \cite{ori_curriculum, c_survey1, c_survey2} is a training strategy that trains a machine learning model in a meaningful order, demonstrating that learning from easy to hard examples can be beneficial.
This strategy has been successfully employed in object detection\cite{c_size, c_svm, c_ohem, c_ijcv, c_1, c_2, c_tpami}. 
\cite{c_size} proposed a curriculum learning strategy to feed training images into the weakly supervised object localization learning loop in an order from images containing bigger objects down to smaller ones.
Based on \cite{c_old}, C-SPCL\cite{c_ijcv} proposed a novel collaborative self-paced curriculum learning framework for weakly supervised object detection, which combined the instance-level confidence inference, the image-level confidence inference, and the self-paced learning mechanism to increase the learning robustness. Under very few annotated samples, \cite{few-example} embedded multi-modal learning and curriculum learning to ameliorate the trade-off between precision and recall in training data production. Different from these methods tackling 2D detection, we explore curriculum learning to 3D detection.

Our proposed method also is inspired by self-paced learning (SPL)~\cite{kumar2010self}, so we provide a brief overview here. Let $\mathcal{D}=\left\{x_{i}, y_{i}\right\}_{i=1}^{N}$ indicates the training set, where $x_i$ and $y_i$ denote the feature and label of sample $i$, respectively. The model $f$ with weight parameters $\phi$ map each $x_i$ to the prediction $f(x_i, \phi)$, and $l_i=L(f(x_i, \phi), y_i)$ denotes the loss, where $L$ is the learning objective. Then the goal is to minimize the loss on the whole training set:
\begin{align} \label{related}
	\min _{\phi ; \boldsymbol{v} \in[0,1]^{N}} \mathbb{E}(\phi, \boldsymbol{v} ; \lambda) = \sum_{i=1}^{N} v_{i} l_{i} - \lambda \sum_{i=1}^{N} v_{i}, 
\end{align}
where $\lambda$ is the age parameter for controlling the learning pace, and $\boldsymbol{v}=\left[v_{1}, v_{2}, \ldots, v_{N}\right]^{\top} \in[0,1]^{N}$ is the latent weight variable. 
The above learning objective is generally optimized with ACS  (Alternative Convex Search)~\cite{c_old}. Concretely, it alternatively optimizes $\phi$ and $\boldsymbol{v}$ while fixing the other. With the fixed $\boldsymbol{v}$, the global optimum  $\phi^*$ can be learned by the existing off-the-shelf supervised learning. Then, with fixed $\phi$, the close-formed solution for $\boldsymbol{v}^*$ can be obtained by solving:
\begin{align} \label{related2}
    v_{i}^{*}=\left\{\begin{array}{lr}
    1, & l_{i}<\lambda, \\
    0, & \text { otherwise. }
    \end{array}\right.
\end{align}
This solution exists an intuitive explanation: if a sample has a loss $l_i$ less than the threshold $\lambda$, then it is taken as an \emph{easy} sample for the current model, and will be selected in training (i.e., $v_i^*=1$). Otherwise, it is \emph{hard} and unselected (i.e., $v_i^*=0$). Initially, $\lambda$ should be set suitable to ensure that a small proportion of easy sample are selected. When the model becomes more mature, $\lambda$ should grow and more harder samples get involved in the training process.

\begin{figure*}[t]
    \centering
    \includegraphics[width=1.0\linewidth]{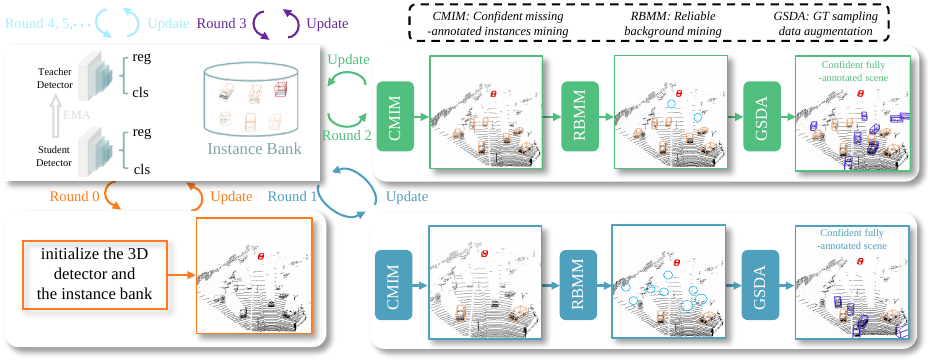}
    \caption{The pipeline of our SS3D++ algorithm. We alternatively improve 3D detector training and confident fully-annotated scene generation in a unified learning scheme. To efficiently train the 3D detector, we construct confident fully-annotated scenes based on the missing-annotated instance mining module, the reliable background mining module , and the GT sampling data augmentation strategy. By leveraging the mutual amelioration between high-quality training scene generation and 3D detector training,  the obtained 3D detector becomes more robust.}    	
    \label{fig:pipline}  
\end{figure*}

\section{Preliminaries}

\subsection{3D Object Detector and Notation}
In this paper, we focus on the 3D object detection in autonomous driving scenarios which aims to predict the attributes of 3D objects from point cloud inputs. 
Formally, we denote the 3D object detector model as $F$, and the 3D object detection task can be formulated as
\begin{align} \label{test}
E_i = F(P_i; \phi),  i = 1, \cdots, M,
\end{align}
where $E_i = \{E_{ij}\}, j= 1, \cdots, N_i$ is a set of detected 3D objects in the $i$-th scene with an input scene $P_i \in \mathbb{R}^{N\times 3}$ containing $N$ points, $\phi$ is the weight parameter of $F$, $M$ is the total number of scenes in the dataset, and 
$E_{ij}$ can be written as
\begin{align}
	E_{ij} = (r, p), r = [x, y, z, l,w,h, \theta], 
\end{align}
where $r$ is the 3D bounding box and $p$ denotes the predicted probability of belonging to the class $c$. In this paper, we consider a 3D bounding box represented in the LiDAR coordinate system, in which $(x, y, z)$ is the 3D center coordinate 
, $(l,w,h)$ is the length, width, and height of a cuboid, respectively, and $\theta$ is the orientation from the bird’s view.

\subsection{Fully-Supervised 3D Object Detection}
Considering the fully-supervised setting, it is often given a training dataset $ \mathcal{D}=\{(P_{1}, G_{1}), \ldots, (P_{M}, G_{M})\} $ and the corresponding ground-truth, where $G_i=\left\{\left( b_{ij}, y_{ij}\right)\right\}, j= 1, \cdots, N_i$, and for each $(b_{ij},y_{ij})$, $b_{ij}$ is the oriented 3D bounding box of the $j$-th object instance in a point cloud scene $P_i$, and $y_{ij} \in \{ 1, \ldots, C \}$ is the class label.
Based on the training dataset $ \mathcal{D}$, the 3D object detector model $F$ can be learned by the following loss function 
\begin{align}
	\mathcal{L}_{F}  =  \mathcal{L}_{cls} + 	\mathcal{L}_{reg}, \label{eq_ori_loss}
\end{align}
where classification loss $ \mathcal{L}_{cls}$ is used to learn the classifier to predict the category of a 3D object, and $ \mathcal{L}_{reg}$ is utilized to learn the location, size and heading angle. 
If we take the proposal-based two-stage 3D detector Voxel-RCNN \cite{voxelrcnn} as an example, the  Eq. (\ref{eq_ori_loss}) degenerates as
\begin{align}
	\mathcal{L}_{F}  =  \mathcal{L}_{cls}^{RPN} + \mathcal{L}_{reg}^{RPN} + \mathcal{L}_{cls}^{Head} + 	\mathcal{L}_{reg}^{Head}, 
\end{align}
where $\mathcal{L}_{cls}^{RPN}$ is the Focal Loss \cite{focal} for classification in the region proposal network stage, $\mathcal{L}_{cls}^{Head}$ is the Binary Cross Entropy Loss for confidence prediction in the detection head stage, and Huber Loss is exploited in $\mathcal{L}_{reg}^{RPN}$ and $\mathcal{L}_{reg}^{Head}$ for box regression.

\section{Sparsely-Supervised 3D Object Detection Framework and Learning Algorithm}
Though fully-supervised 3D object detection obtains impressive performance, they require large-scale, precisely densely-annotated 3D training data. This data collection process is quite time-consuming and labor-intensive, which prevents the potential applicability of 3D object detection in real-world settings.
In this work, we study the following sparsely annotated 3D object detection setting, i.e., we have only access to one 3D bounding box annotation $\tilde{G}_{i}=\{(b_{i1},y_{i1})\}$ for any scene $P_i$, and the annotations of remaining object instances are missing. The subsequent experiments verifies that the proposed method is robust in terms of the way of object selection. Such inexact and incomplete supervision
greatly saves annotation efforts. In contrast to prior weakly-supervised attempts \cite{fgr,weaklysuper,3dioumatch} (as shown in Fig. \ref{fig:diffierent}), our novel sparse supervision form introduces more potential practical and commercial benefits, while bringing some new challenges (please refer to Section \ref{sec:introduction}). In this section, we will present the proposed SS3D++ method to address these challenges.

\subsection{Overview SS3D++ Framework and Modeling}
We aim to facilitate the learning strategy of a 3D detector to obtain the optimal detection performance when training from scratch on the sparsely annotated dataset $ \tilde{\mathcal{D}}=\{(P_{1}, \tilde{G}_{1}), \ldots, (P_{M}, \tilde{G}_{M})\}$.
However, such sparsely annotated data provides inexact and incomplete supervision, inevitably raising the severe deterioration of fully-supervised 3D detectors, as shown in Tables \ref{tab:tab1} and \ref{tab:tab1waymo}.

To reduce the negative influence of inexact and incomplete supervised information, it is encouraged to generate pseudo annotations for those missing-annotated instances.
Without loss of generality, we consider the formulation for the $i$-th scene, which can be easily applied to all training scenes by cancelling out the index $i$.
Note that the use of $i$ is for the easy clarity, the actual optimization is carried out through minibatch.
To mine more confident missing-annotated positive samples, we propose to use a meaningful multi-criteria sample selection curriculum  (Eqs. (\ref{Equ}), (\ref{Eqv}), and (\ref{Eqk})), and further improve the 3D detector using the selected positive instances in Eq. (\ref{eqmodel}). 
To sum up, our approach can be formulated as the following optimization objective:
\begin{align}
 \   \mathbf{u}_{i}^*, \mathbf{v}_{i}^*, \mathbf{k}_{i}^* &= \mathop{\arg\min}_{\mathbf{u}_{i}, \mathbf{v}_{i}, \mathbf{k}_{i}} \mathcal{E}(\mathbf{u}_{i}, \mathbf{v}_{i}, \mathbf{k}_{i}; \phi^*, \lambda) \nonumber \\
  & \triangleq  \mathcal{L}_{c\lambda} + \mathcal{L}_{r\lambda} + \mathcal{L}_{d\lambda},    
\end{align}
\begin{align}
    \ & 	\phi^*,\tilde{U}_{i}^{*} = \mathop{\arg\min}_{\phi, \tilde{U}_{i}} \mathcal{E}(\phi, \tilde{U}_{i}; \mathbf{u}_{i}^*, \mathbf{v}_{i}^*, \mathbf{k}_{i}^*) \triangleq  \mathcal{L}_s + \mathcal{L}_p,
    \label{eqmodel}
\end{align}
where 
\begin{align}
 \ & \mathcal{L}_{c\lambda} = \sum_{j=2}^{N_i}\sum_{c=1}^{C} u_{ijc} \mathcal{L}^c_{cls}(p_{ij} (\phi^*), \tilde{y}_{ij}^{*}) -\sum_{j=2}^{N_i}\sum_{c=1}^{C} \lambda^u_{ic} u_{ijc},  \label{Equ} \\ 
	\ &  \mathcal{L}_{r\lambda} =  \sum_{j=2}^{N_i}\sum_{c=1}^{C} v_{ijc} \mathcal{L}^c_{reg}(\mathcal{A}(r_{ij}(\phi^*)), \tilde{r}_{ij}) -  \sum_{j=2}^{N_i} \sum_{c=1}^{C}\lambda^v_{ic} v_{ijc},  \label{Eqv} \\
	\ &  \mathcal{L}_{d\lambda} = \sum_{j=2}^{N_i}\sum_{c=1}^{C} k_{ijc} d^c(r_{ij}(\phi^*)) -  \sum_{j=2}^{N_i} \sum_{c=1}^{C} \lambda^k_{ic} k_{ijc},  \label{Eqk}
\end{align}
are the loss for sparse annotation and pseudo instances, respectively, 
$\tilde{G}_{i1} = (b_{i1},y_{i1})$ is the given sparse supervision of the $i$-th scene, $\tilde{U}_{ij} = (\tilde{b}_{ij},\tilde{y}_{ij})$ and $\tilde{U}_{i}^{*}$ represent the previously generated pseudo instances and the currently generated pseudo instances, respectively, while $\tilde{b}_{ij}$ and $\tilde{y}_{ij}$ are bounding boxes and class prediction for the mined $j$-th positive instance, respectively. $\phi$ is the weight parameter of detector $F$, which can be instantiated as any existing off-the-shelf fully-supervised 3D object detectors (please see Section \ref{experiment}). 
$u_{ijc}, v_{ijc}, k_{ijc}\in \{0,1\}$ are the learnable sample selection variables encoding whether the detected result in the $i$-th scene are determined as reliable positive instances to further train the detector, 
and $\mathbf{u}_{i}, \mathbf{v}_{i}, \mathbf{k}_{i}$ are weight matrix, $\mathbf{u}_{i}, \mathbf{v}_{i}, \mathbf{k}_{i} \in \{0, 1\}^{(N_i-1) \times C}$, denoting the collection of the selection variables $u_{ijc}, v_{ijc}, k_{ijc}$ for all mined positive instances with different class $c$. $\lambda^u_{ic},\lambda^v_{ic},\lambda^k_{ic}$ are the hyperparameter for the SPL regularization term \cite{kumar2010self,meng2017theoretical}, which enables the possible selection of high-confidence images during optimization.
$r_{ij},\tilde{r}_{ij}$ are the predicted bounding boxes of the detector $F$ by inputting scenes $P_i$ and $\mathcal{A}(P_i)$, respectively, where $\mathcal{A}(\cdot)$ is a set of global augmentation strategies including random rotation, flipping, and scaling.
$d(r_{ij}(\phi))$ represents the reciprocal of the density estimation of the predicted bounding boxes $r_{ij}$ by the detector $F(\cdot, (\phi))$, which is the cardinality of $S_{ij}$, i.e., the number of the points inside the 3D bounding box.

Actually, above formation is a bi-level optimization objective \footnote{We drop explicit dependence of $\phi$ on $\mathbf{u}_i, \mathbf{v}_i, \mathbf{k}_i$ for brevity here.} \cite{liu2021investigating}.
Calculating the optimal detector $F$ in the inner-loop can be very expensive. This motivates us to adopt online iterative updation by the sequence $\phi, \tilde{U}_{i},\mathbf{u}_i, \mathbf{v}_i, \mathbf{k}_i$, until the maximum iteration round number is reached. 
Next we show how to solve each parameter as follows.

 \textbf{Update $\mathbf{u}_{i}, \mathbf{v}_{i}, \mathbf{k}_{i}$.}
For each training iteration, we can obtain $\mathbf{u}_{i}, \mathbf{v}_{i}, \mathbf{k}_{i}$ for the $i$-th scene by solving Eqs. (\ref{Equ}), (\ref{Eqv}), and (\ref{Eqk})  based on last step $\phi^{(t)}$ and $\tilde{U}_{i}^{(t)}$. Then the closed-form solution is
\begin{align}
u^{(t+1)}_{ijc} & =  \left\{
\begin{array}{ll}
	1  &  \mathcal{L}^c_{cls}(p_{ij}, y_{ij}^{(t)};\phi^{(t)}) < \lambda^u_{ic} \\
	0  & \mathcal{L}^c_{cls}(p_{ij}, y_{ij}^{(t)};\phi^{(t)}) \geq \lambda^u_{ic}
\end{array}, \right.   \label{equu}\\ 
	v^{(t+1)}_{ijc} &=  \left\{
	\begin{array}{ll}
		1  &  \mathcal{L}^c_{reg}(\mathcal{A}(r_{ij}), \tilde{r}_{ij};\phi^{(t)}) < \lambda^v_{ic} \\
		0  & \mathcal{L}^c_{reg}(\mathcal{A}(r_{ij}), \tilde{r}_{ij};\phi^{(t)}) \geq \lambda^v_{ic}
	\end{array}, \right.   \label{eqvv}\\ 
k^{(t+1)}_{ijc} &=  \left\{
\begin{array}{ll}
1  &  d^c(r_{ij};\phi^{(t)}) < \lambda^k_{ic} \\
0  & d^c(r_{ij};\phi^{(t)}) \geq \lambda^k_{ic}
\end{array}, \right.  \label{eqkk}
\end{align}
and the implementation details of this multi-criteria sample selection process can be refer to Section \ref{foreground}.

\textbf{Update $\tilde{U}_{i}$.} The next step is to update pseudo instances $\tilde{U}_{i}$ of the training scene $P_i$ by solving the following minimization sub-problem:
\begin{align} \label{pseudo}
	\tilde{U}_{i}^{*} = \mathop{\arg\min}_{\tilde{U}_{i}} \mathcal{L}_p.
\end{align}
It is easy to see that the global optimum of the above problem can be obtained by directly setting the pseudo annotations equal to the predictions of the detector. Here, we adopt teacher network with EMA predictions for each of the training instances to generate high quality of pseudo annotations $\tilde{U}_{i}^{(t+1)}$, which is the commonly used technique in semi-supervised learning \cite{tarvainen2017mean}.

\renewcommand{\algorithmicrequire}{ \textbf{Input:}} 
\renewcommand{\algorithmicensure}{ \textbf{Output:}} 
\begin{algorithm}[t]
	\caption{Our SS3D++ Method.}
	\label{alg:Framwork}
	\begin{algorithmic}[1] 
		\REQUIRE Teacher detector $F_m$, student detector $F_s$, total rounds $\mathcal{T}$, instance bank ${\mathcal{B}}$, sparsely annotated dataset $\tilde{\mathcal{D}}$, global augmentation strategies $\mathcal{A}$;\\
		\STATE Initialize the instance bank ${\mathcal{B}}^{(0)}$ with given sparse annotations;
		\STATE Pre-train $F_m$ with dataset $\tilde{\mathcal{D}}$ via Eq. (\ref{eq_ori_loss}), and initialize the $F_s$ with the same pre-trained weight parameters;
		\STATE Get initial density vector $d_0$ via Eq. (\ref{con:init}) in Section \ref{adaptive};
		\FOR { $t=1$ to $\mathcal{T}$ }
		\IF{$t$ \textgreater $1$}
        \STATE Perform multi-criteria sample selection curriculums via Eqs. (\ref{equu}), (\ref{eqvv}), and (\ref{eqkk}); 
        \STATE Using $F_m$ to generate pseudo instances $\tilde{U}_{i}$ via Eq. (\ref{pseudo}) and updating instance bank ${\mathcal{B}^{(t)}}$;
		\ENDIF
		\STATE Perform reliable background mining process (Algorithm \ref{alg:mimm}) via the $F_m$ to obtain broken scene set $\tilde{P}^{(t)}$ with reliable background;
		\WHILE{detector not converge}		
		\STATE Perform confident fully-annotated scene generation to get $\tilde{\mathcal{D}_c}^{(t)}$ based on $\tilde{P}^{(t)}$ and ${\mathcal{B}^{(t)}}$;
		\STATE Train $F_s$ via Eq. (\ref{student}) based on $\tilde{\mathcal{D}_c}^{(t)}$; 
        \STATE Update $F_m$ via Eq. (\ref{teacher});
		\ENDWHILE
		\ENDFOR
		\ENSURE Learned student detector $F_s$. \\ 
	\end{algorithmic}
\end{algorithm}

\begin{figure*}[t]
	\centering
	\includegraphics[width=\linewidth]{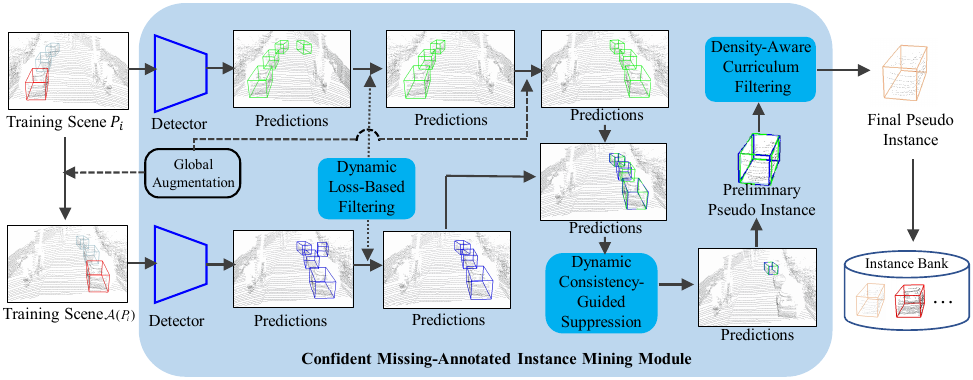}
	\caption{
		Illustration of proposed confident missing-annotated instance mining module.
		The scene $P_i$ and the corresponding augmented scene $\mathcal{A}({P}_i)$ are input into the 3D detector.
		Then we leverage the dynamic loss-based filtering to remove the predictions of $P_i$ and $\mathcal{A}({P}_i)$ with a low confidence score (i.e., large classification loss).
		Further, the dynamic consistency-guided suppression is proposed to filter out low-quality predictions.
		Lastly, we carry out the density-aware curriculum filtering process and store the remaining predictions into the instance bank as confident pseudo instances.
	}
	\label{fig:mimm}  
\end{figure*}

\textbf{Update $\phi$.} Finally, we train the 3D object detector given $\mathbf{u}^{(t+1)}_i, \mathbf{v}^{(t+1)}_i, \mathbf{k}^{(t+1)}_i$ and $\tilde{U}_{i}^{(t+1)}$. 
Specifically, after obtaining pseudo instances from the teacher detector, we can formulate the loss of student detector as follows:
\begin{align} \label{student_loss}
    \ & {\phi}_{s}^{*} = \mathop{\arg\min}_{{\phi}} (\mathcal{L}_s + \mathcal{L}_p).
\end{align}
Then, the the weight parameter $\phi_{s}$ of the student detector is updated via stochastic gradient descent: 
\begin{align} \label{student}
\ &	\phi_{s} \leftarrow \phi_{s} - \gamma \frac{\partial (\mathcal{L}_s+\mathcal{L}_p)}{\partial \phi_{s}},
\end{align}
where $\gamma$ is the learning rate.
Next, we apply the effective EMA~\cite{tarvainen2017mean} approach which has been shown to be effective in many existing works to update the learned weight $\phi_{m}$ of the teacher detector as follows:
\begin{align} \label{teacher}
	\phi_{m} \leftarrow \alpha \phi_{m}+(1-\alpha) \phi_{s}, 
\end{align}
where $\alpha$ is the EMA coefficient.
Note that the training data are the union set of initial sparse annotated instances and selected instances ($u_{ij}=v_{ij}=k_{ij}=1$) with the pseudo annotation $\tilde{U}_{i}$. To increase the learning efficiency of the detector, we further generate confident fully-annotated scenes based on the mined reliable background and confident missing-annotated instances. 
The details can refer to Sec. \ref{scene}.
Thus, this step can be solved by the existing standard off-the-shelf 3D object detector, described as PointRCNN \cite{pointrcnn}, Part-A$^2$\cite{parta2}, PV-RCNN\cite{pvrcnn}, Voxel-RCNN\cite{voxelrcnn}, etc.

With the above analysis, it is clear to see that 
our approach aims to automatically select confident instances for generating pseudo instances (i.e., Eqs. (\ref{equu}), (\ref{eqvv}), and (\ref{eqkk})) , and update the weight parameter $\phi$ in the above formulation with an iteratively-learning manner.
The overall algorithm of our SS3D++ algorithm for sparsely-supervised 3D object detection framework is shown in Algorithm \ref{alg:Framwork}.
Given a 3D teacher detector, initially, we train the detector from the scratch on the sparsely annotated dataset $\tilde{\mathcal{D}}$ (i.e., step 2).
To increase the diversity of training data and eliminate the negative impact of missing-annotated instances, we leverage the teacher detector to mine reliable missing-annotated instances in a meaningful curriculum learning order through the missing-annotated instance mining module with strict multi-criteria sample selection curriculums through Eqs. (\ref{equu}), (\ref{eqvv}), and (\ref{eqkk}) (step 6, refer to Algorithm \ref{alg:rbmm} and Fig. \ref{fig:mimm}).
Furthermore, we add the mined high quality missing-annotated instances into the instance bank. 
Then, we use the teacher detector to get the broken scene with mined reliable background through the reliable background mining module (details can refer to Algorithm \ref{alg:mimm} and Fig. \ref{fig:nsrm}),
which aims to further prevent missing-annotated instances and the region near those instances from being mistaken as background. 
To efficiently train the student detector with sufficient knowledge, we further construct confident fully-annotated scenes based on the mined reliable background and confident missing-annotated instances (step 11, refer Section \ref{scene} for details).  
By this iteratively learning style of reliable background mining, confident missing-annotated instances mining, confident fully-annotated scene generation, and detector updating through Eqs. (\ref{student}) and (\ref{teacher}), more confident fully-annotated scenes are obtained, and the teacher/student detector becomes more robust and well-performing.

\subsection{Confident Missing-Annotated Instances Mining} \label{foreground}

In this section, we detailedly illustrate the multi-criteria sample selection curriculums in Eqs. (\ref{equu}), (\ref{eqvv}), and (\ref{eqkk}). We implement these curriculums via a confident missing-annotated instance mining module, shown in Fig. \ref{fig:mimm}, to effectively mine missing-annotated instances for each scene.
Specifically, the collectively choosing process of confident instances is influenced by three critical mechanisms: dynamic loss-based filtering, dynamic consistency-guided suppression, and density-aware curriculum filtering.
Then the updating equation of the pseudo instances can be formulated by updating $\mathbf{u}_i$, $\mathbf{v}_i$, and $\mathbf{k}_i$ along the iteratively learning process.

Next, we will delve into the detailed explanation of the multi-criteria sample selection process and corresponding adaptive sample selection curriculum setups, thereby fundamentally comprehending the origins of the three variables: $\mathbf{u}_i$, $\mathbf{v}_i$ and $\mathbf{k}_i$.

\begin{figure}[t]
    \centering
    \includegraphics[width=\linewidth]{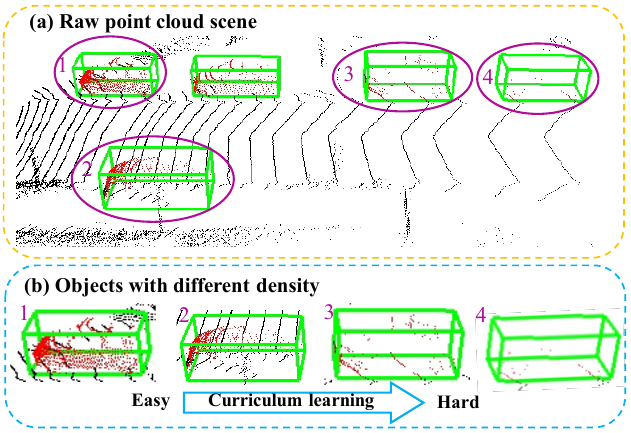}
    \caption{(a) Illustration of learning hardness (i.e., density) of different 3D objects on the KITTI dataset. (b) We select mined 3D objects in a meaningful order from ``easy'' to ``hard''. The red represents the foreground points and the green represents the ground-truth bounding boxes. 	 
    }
    \label{fig:density}  
\end{figure}

\subsubsection{Multi-Criteria Sample Selection Process} \label{criteria}
As shown in Fig. \ref{fig:mimm}, it can be seen that three selection variables $\mathbf{u}_i, \mathbf{v}_i, \mathbf{k}_i$ play distinct roles, which correspond to the following three-criteria sample selection curriculums. 
\begin{itemize}
	\item \textbf{Dynamic loss-based filtering.} Eq. (\ref{equu}) tries to filter out samples with large classification loss, which means that we tend to select instances with high-confident prediction. Along with the amelioration of detector, more high-confident prediction gradually involves in training to further boost the detector. 
	\item \textbf{Dynamic consistency-guided suppression.} Inspired by the consistency-based semi-supervised learning~\cite{fixmatch}, we further require that the generated bounding boxes should have fine consistency property. Specifically, we compute the mismatch between the prediction $\tilde{r}_{ij}$ of the augmented input scene and the augmented prediction $\mathcal{A}(r_{ij})$ under a same global augmentation strategy $\mathcal{A}$. As demonstrated in Eq. (\ref{eqvv}), we suppress the effect of low consistent predictions (i.e., large regression loss) during the training process, which helps to reduce the effect of noisy bounding boxes, and thus improve the performance of the detector.
	\item \textbf{Density-aware curriculum filtering.} 	
     As shown in Fig. \ref{fig:density}, due to the occlusion, truncation and the distance from the LiDAR sensor, the objects possess different densities. The lower density of objects may contain less spatial information, which may make detector hard to extract fine features at the initial training stage. 	 
	To this goal, we mine the confident objects by considering the hardness of learning different objects. Specifically, we use Eq. (\ref{eqkk}) to progressively select mined instances from easier objects (high density) to harder objects (low density) during the training process. The density $d(r_{ij}(\phi^{(t)}))$ of generated bounding boxes $r_{ij}$ can be computed using the cardinality of $S_{ij}$.
 Through controlling the instances involving in training from  ``easy'' to ``hard'' order in terms of the density, we can obtain a more robust detector with these newly generated annotated training data.
\end{itemize}

\subsubsection{Adaptive Sample Selection Curriculum Setups} \label{adaptive}
Note that hyperparameters $\lambda^u_{ic},\lambda^v_{ic},\lambda^k_{ic}$ in Eqs. (\ref{equu}), (\ref{eqvv}), and (\ref{eqkk}) are related to how many instances of each class are used to train the detector. We propose the following curriculum setups to adaptively determine appropriate hyperparameters $\lambda^u_{ic},\lambda^v_{ic},\lambda^k_{ic}$ to guarantee that the instances involving in the training pool can robustly increase the detector's performance during the training process.
\begin{itemize}
	\item \textbf{Adaptive loss curriculum setup.} Notice that the range of loss predictions varies greatly for different detectors. To overcome this, we use the histogram of classification loss distribution to adaptively determine the proper $\lambda^u_{ic}$ during the training process. Specifically, we first compute the classification loss $\mathcal{L}_{cls}(p_{i\cdot}(\phi^{(t)}), y_{i\cdot}^{(t)})$ in Eq. (\ref{equu}) of all candidate bounding boxes based on the current detector $F(\cdot; \phi^{(t)})$. Then we calculate the histogram of the loss distribution,
	and then find the point with the fastest frequency decay rate of adjacent bins. We then set the parameter $\lambda^u_{ic}$ as this breakpoint (marked by the blue arrow), as shown in Fig. \ref{fig:his}(a). Finally, we select the bounding boxes with classification loss values smaller than this breakpoint involving in training.
	
	\item \textbf{Adaptive consistency curriculum setup.} We adopt similar curriculum setup as the loss-base one. As shown in Fig. \ref{fig:his}(b), we first calculate the regression loss $\mathcal{L}_{reg}(\mathcal{A}(r_{ij}(\phi^{(t)})), \tilde{r}_{ij})$ in Eq. (\ref{eqvv}), and then find the breakpoint (marked by the blue arrow) based on the histogram of the regression loss distribution to set $\lambda^v_{ic}$. Bounding boxes with regression loss values larger than this breakpoint are removed.

\begin{figure}[t]
    \centering
    \includegraphics[width=\linewidth]{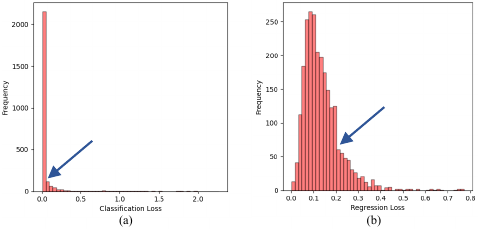}
    \caption{
    Setting of dynamic threshold based on the histogram. 
    (a) and (b) show the histogram of confidence score group and paired IoU group, respectively.
    The blue arrow points correspond to the steepest descent point, i.e., the selection of the dynamic threshold.
    }
    \label{fig:his} 
\end{figure}

	\item \textbf{Adaptive density curriculum setup.} Different datasets are collected by different quality of LiDAR, resulting in drastic density variations in the point clouds. To enable the model to adapt to different datasets, we propose to use the following linear curriculum function to set $\lambda^k_{ic}$: 
		\begin{equation}
		\label{con:linear}
		\lambda^k_{ic}(t) = \max (d_{min}^c, d_{0}^c-\frac{d_{0}^c - d_{min}^c}{\mathcal{T}_{\text {down}}} \cdot t),
	\end{equation}
where the initial density value $d_0^c$ for class $c$ of a training scene can be written as
	\begin{equation}
		\label{con:init}
		d_0^c = \frac{1}{\sum_{i=1}^{M} N_i} \sum_{i=1}^{M} \sum_{j=1}^{N_i} d(r_{ij}^c (\phi^{(0)})), c = 1,\cdots,C.
	\end{equation}
We use $d_0=(d_0^1, \cdots,d_0^C)$ to denote initial density value vector for all classes. Here, $\phi^{(0)}$ represents the model weight of the detector trained directly on the sparsely annotated dataset $\tilde{D}$, $r_{ij}^c$ is the bounding box generated by the detector $F(\cdot; \phi^{(0)})$. Besides, $d_{min}^c$ denotes the minimum density of the hardest instances, 
and ${\mathcal{T}_{\text {down}}}$ denotes the preset training iteration steps (We set it as the four-fifth of the total training steps). By this formulation, different classes are separately handled by considering their inherently different densities, and we filter out the bounding box predictions whose densities are less than $\lambda^k_{ic}(t)$ for class $c$ at the $t$-th training round.
\end{itemize}

\begin{figure*}[t]
	\centering
	\includegraphics[width=\linewidth]{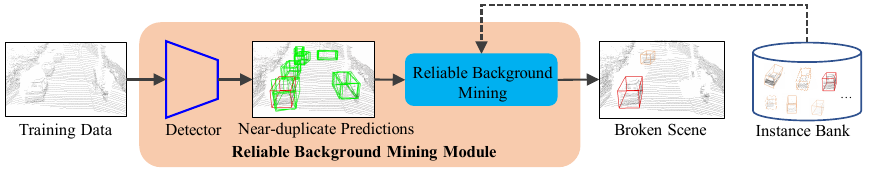}
	\caption{
		Illustration of the proposed reliable background mining module. 
		To start, we feed the original point cloud to the detector without NMS to produce near-duplicate predictions, and leverage the instances stored in the instance bank to filter out unreliable 3D object points.  
		This will lead to broken scene with reliable background, which is further processed through the GT sampling data augmentation in Sec. \ref{scene}. 
	}
	\label{fig:nsrm} 
\end{figure*}

The illustration of the confident missing-annotated instance mining process can be found in \cref{fig:mimm}, and the algorithm for this process is shown in \cref{alg:rbmm}.
The final selected pseudo instances are stored in the instance bank $\mathcal{B}$, with their point cloud inside the 3D bounding boxes and predicted class labels. By these sample selection curriculums, when the detector becomes more mature (i.e., in the subsequent iterations), the negative influence of missing-annotated instances gradually decreases and our instance bank can store more and more confident positive instances to further support the reliable background mining process.
Noted that we do not perform pseudo instance mining in the first iteration, since many missing-annotated instances bring poor detection performance at the early training stage.

\textbf{Remark.} Different from traditional SPL \cite{kumar2010self,meng2017theoretical} modelling selection variables and model parameter in the same-level optimization problem, we formulate selection variables and model parameter as outer- and inner-level optimization, respectively, which is a typical bi-level optimization problem \cite{liu2021investigating,franceschi2018bilevel}. 
More recently, several works attempt to learn selection variables using an unbiased meta data \cite{shu2019meta,shu2022cmw,shu2023cmw,shu2020meta} in a meta-learning manner. 
Comparatively, we regard the well-studied SPL model as meta-knowledge \cite{shu2023dac} to help promote better multi-criteria sample selection process. More importantly, we use multi-criteria sample selection curriculums to jointly mine confident missing-annotated instances, which is more effectively and robustly to filter out unreliable 3D object instances compared with previous single-criteria methods \cite{c_tpami,few-example}.

Though the confident missing-annotated instances mining module mines some high-quality positive instances, there are still omissions, which naturally lead to the detector overfitting to the unreliable background (the missing-annotated instances and the region near those instances may be incorrectly marked as background).
Hence, we propose a reliable background mining module to possibly mine reliable background.

\renewcommand{\algorithmicrequire}{ \textbf{Input:}} 
\renewcommand{\algorithmicensure}{ \textbf{Output:}} 
\begin{algorithm}[t]
\caption{Confident Missing-Annotated Instance Mining.}
\label{alg:rbmm}
\begin{algorithmic}[1] 
\REQUIRE Teacher detector $F_m$ with weight parameter $\phi^{(t)}$ at round $t$, instance bank ${\mathcal{B}^{(t)}}$, sparsely annotated dataset $\tilde{\mathcal{D}}$, initial density $d_0$, global augmentation strategies $\mathcal{A}$;\\
    \FOR{ $P_i$ in $\tilde{\mathcal{D}}$ }
        \STATE Obtain $\hat{P}_i = \mathcal{A}({P}_i)$ by applying global augmentation strategies $\mathcal{A}$ to original scene $P_i$;
        \STATE Obtain ${E}_i$ and $\hat{{E}}_i$ by inputting $P_i$ and $\hat{P}_i$ to the detector $F_m(\cdot; \phi^{(t)})$ based on Eq. (\ref{test}); 
        \STATE Set $\lambda^u_{ic}, \lambda^v_{ic},\lambda^v_{ic}$ by adaptive sample selection curriculum setups in Eq. (\ref{adaptive});
        \STATE Perform joint multi-criteria sample selection process via Eq. (\ref{equu}), Eq. (\ref{eqvv}) and Eq. (\ref{eqkk});
        \STATE Generate high quality of pseudo annotations $\tilde{U}_{i}$ for selected instances via Eq. (\ref{pseudo});
        \STATE Update instance bank $\mathcal{B}_i^{(t)}$ by adding newly selected confident missing-annotated instances, denoted by $\mathcal{B}_i^{(t+1)}$.
    \ENDFOR 
\ENSURE Updated instance bank $\mathcal{B}_i^{(t+1)}$\\ 
\end{algorithmic}
\end{algorithm}

\begin{figure}[t]
    \centering
    \includegraphics[width=6cm]{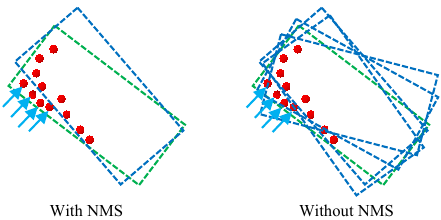}
    \caption{Illustration of detection results with NMS (left) and without NMS (right), where the green and blue dashed boxes represent the ground truth and predicted bounding box, respectively.
    To handle the omission of foreground points (red points pointed by blue arrows) due to inaccurate location boxes, we remove the NMS operation from the detector. 
    }
    \label{fig:nms} 
\end{figure}

\subsection{Reliable Background Mining} \label{background}
In this section, the principles and mechanisms of step 9 in Algorithm \ref{alg:Framwork} are elaborated in detail.
To reduce the possibility of wrongly labeling positive instances as background, it is encouraged to mine potential foreground points as far as possible, and then the remains should be the reliable background. To this goal, we propose the reliable background mining module shown in Fig. \ref{fig:nsrm}. Specifically, we firstly set a low confidence score threshold for the detector.
Then, to eliminate the foreground points omissions issue caused by inaccurate location boxes, frequently occurring at the initial training phase of the detector, we remove the Non-Maximum Suppression (NMS) operation \cite{nms1, nms2}
to get near-duplicate prediction results as illustrated in Fig. \ref{fig:nms}.
Those two simple strategies can guarantee that the produced results contain potential foreground points as far as possible, and the detailed recall rate analysis can be seen in Fig. \ref{fig:three_fig} (a). 

As illustrated in Fig. \ref{fig:nsrm}, we construct an instance bank which initially just stores the sparsely annotated instances and then progressively stores mined highly confident pseudo instances (please see in Section \ref{foreground}).  

Specifically, we denote the predicted 3D bounding boxes as $R_i$, and the bounding boxes of the scene $P_i$ in instance bank $\mathcal{B}$ as $\mathcal{B}_i$, which contains the sparsely annotated instance and previously mined pseudo instances. 
And then we delete all points of the predicted bounding boxes $R_i$ in the scene $P_i$ to get the confident scene $\tilde{P}_{i}$.
Next, the bounding boxes and corresponding points stored in the instance bank $\mathcal{B}_i$ of the $i$-th scene will fill into the scene $P_i$.
The instance bank helps to avoid wrongly discarding reliable 3D objects whilst removing the ambiguous points for each training scene. 

The whole pipeline is illustrated in Fig. \ref{fig:nsrm}, and the learning process is presented in Algorithm \ref{alg:mimm}. The final obtained broken scenes via Algorithm \ref{alg:mimm} contain reliable background and some confident foreground objects.

\renewcommand{\algorithmicrequire}{ \textbf{Input:}} 
\renewcommand{\algorithmicensure}{ \textbf{Output:}} 
\begin{algorithm}[t]
	\caption{Reliable Background Mining.}
	\label{alg:mimm}
	\begin{algorithmic}[1] 
		\REQUIRE Teacher detector $F_m$ with weight parameter $\phi^{(t)}$ at round $t$, instance bank ${\mathcal{B}^{(t)}}$, sparsely annotated dataset $\tilde{\mathcal{D}}$, low score threshold $\tau_{low}$;\\
		\STATE Initialize empty scene set $\tilde{P}^{(t)}$;
		\FOR{ $P_i$ in $\tilde{\mathcal{D}}$ }
		\STATE Obtain ${E}_i$ via Eq. (\ref{test}) by setting the detector $F_m$ w/o NMS operation and w/ threshold $\tau_{low}$;
    	\STATE Fetch 3D objects ${\mathcal{B}_i}$ from instance bank          $\mathcal{B}^{(t)}$;
        \STATE Delete all points of ${E}_i$ for the scene $P_i$;
		\STATE Generate the confident scene $\tilde{P}_i^{(t)}$ by filling all points of ${\mathcal{B}_i}$ for the scene $P_i$;
		\STATE $\tilde{P}^{(t+1)}$ = $\tilde{P}^{(t)} \cup \tilde{P}_i^{(t)}$;
		\ENDFOR
		\ENSURE Broken scenes set $\tilde{P}^{(t+1)}$ with reliable background. \\ 
	\end{algorithmic}
\end{algorithm}

\subsection{Confident Fully-Annotated Scene Generation} \label{scene}
While reliable background mining eliminates most of the ambiguity in supervision, the point cloud scene may only contain a few instances of ground truth, and the distribution of these instances may differ significantly from that of real scenes. This can significantly degrade the network's performance if they are used directly for training.
Following the idea of GT-Sampling \cite{second} data augmentation, we propose to generate confident fully-annotated scenes by combining mined positive instances from constantly updated instance bank and generating broken scenes with reliable background, which enables the the 3D detector to learn from high-quality fully-annotated scenes supervision signals under challenging sparsely-annotated setting.
Now, the training process of teacher and student detector can be obtained via Eqs. (\ref{student}) and (\ref{teacher}).

\begin{table*}[t]  %
\caption{Comparison with different detectors trained with full annotations and extremely sparse split (20\% instances of full annotations) on KITTI \emph{val} split, where green numbers represent decrease, red numbers represent increase. We report the mAP with 40 recall positions.
} 
\resizebox{\textwidth}{!}{
	\begin{tabular}{c|c||ccc|ccc|ccc|ccc|ccc|ccc} 
\hline
\multirow{2}{*}{Method} & \multirow{2}{*}{Data} & \multicolumn{3}{c|}{3D Detection (Car)} & \multicolumn{3}{c|}{BEV Detection (Car)} & \multicolumn{3}{c|}{3D Detection (Ped.)} & \multicolumn{3}{c|}{BEV Detection (Ped.)} & \multicolumn{3}{c|}{3D Detection (Cyc.)} & \multicolumn{3}{c}{BEV Detection (Cyc.)} \\ 
 &  & Easy & Mod & Hard & Easy & Mod & Hard & Easy & Mod & Hard & Easy & Mod & Hard & Easy & Mod & Hard & Easy & Mod & Hard \\ 
\hline
1. PointRCNN\cite{pointrcnn} & Full & 89.66 & 80.59 & 78.05 & 93.19 & 89.13 & 86.84 & 61.58 & 54.58 & 47.93 & 66.27 & 58.26 & 51.57 & 89.03 & 70.61 & \multicolumn{1}{c|}{66.34} & 91.99 & 71.90 & 69.53 \\
2. PointRCNN\cite{pointrcnn} & Sparse (20\%) & 64.33 & 54.91 & 53.97 & 77.50 & 71.31 & 70.84 & 39.35 & 32.41 & 28.40 & 42.14 & 36.13 & 31.03 & 60.46 & 46.95 & \multicolumn{1}{c|}{43.45} & 63.17 & 48.34 & 44.61 \\
3. SS3D++ (Ours) & Sparse (20\%) & 91.80 & 79.85 & 77.65 & 95.83 & 88.37 & 86.48 & 63.95 & 55.98 & 48.13 & 67.25 & 59.98 & 51.39 & 89.84 & 74.13 & \multicolumn{1}{c|}{70.51} & 92.87 & 77.30 & 72.65 \\
4. Improved 2 $\rightarrow$ 1 & - & {\color[HTML]{008F0A} -25.33} & {\color[HTML]{008F0A} -25.68} & {\color[HTML]{008F0A} -24.08} & {\color[HTML]{008F0A} -15.69} & {\color[HTML]{008F0A} -17.82} & {\color[HTML]{008F0A} -16.00} & {\color[HTML]{008F0A} -22.23} & {\color[HTML]{008F0A} -22.17} & {\color[HTML]{008F0A} -19.53} & {\color[HTML]{008F0A} -24.13} & {\color[HTML]{008F0A} -22.13} & {\color[HTML]{008F0A} -20.54} & {\color[HTML]{008F0A} -28.57} & {\color[HTML]{008F0A} -23.66} & \multicolumn{1}{c|}{{\color[HTML]{008F0A} -22.89}} & {\color[HTML]{008F0A} -28.82} & {\color[HTML]{008F0A} -23.56} & {\color[HTML]{008F0A} -24.92} \\
5. Improved 3 $\rightarrow$ 1 & - & {\color[HTML]{FF0000} +2.14} & {\color[HTML]{008F0A} -0.74} & {\color[HTML]{008F0A} -0.40} & {\color[HTML]{FF0000} +2.64} & {\color[HTML]{008F0A} -0.76} & {\color[HTML]{008F0A} -0.36} & {\color[HTML]{FF0000} +2.37} & {\color[HTML]{FF0000} +1.40} & {\color[HTML]{FF0000} +0.20} & {\color[HTML]{FF0000} +0.98} & {\color[HTML]{FF0000} +1.72} & {\color[HTML]{008F0A} -0.18} & {\color[HTML]{FF0000} +0.81} & {\color[HTML]{FF0000} +3.52} & \multicolumn{1}{c|}{{\color[HTML]{FF0000} +4.17}} & {\color[HTML]{FF0000} +0.88} & {\color[HTML]{FF0000} +5.40} & {\color[HTML]{FF0000} +3.12} \\ \hline
1. Part-A$^2$\cite{parta2} & Full & 92.15 & 82.91 & 81.99 & 92.90 & 90.06 & 88.35 & 66.88 & 59.67 & 54.62 & 70.53 & 64.19 & 59.24 & 90.34 & 70.13 & \multicolumn{1}{c|}{66.92} & 91.95 & 74.63 & 70.63 \\
2. Part-A$^2$\cite{parta2} & Sparse (20\%) & 78.54 & 68.81 & 67.03 & 85.05 & 79.63 & 78.03 & 45.73 & 41.37 & 38.40 & 50.32 & 45.95 & 43.17 & 73.61 & 55.10 & \multicolumn{1}{c|}{50.76} & 75.09 & 57.77 & 53.26 \\
3. SS3D++ (Ours) & Sparse (20\%) & 92.74 & 82.19 & 81.11 & 95.93 & 89.32 & 87.83 & 68.86 & 62.40 & 57.26 & 71.51 & 65.80 & 60.12 & 92.35 & 74.77 & \multicolumn{1}{c|}{70.87} & 93.46 & 75.75 & 71.60 \\
4. Improved 2 $\rightarrow$ 1 & - & {\color[HTML]{008F0A} -13.61} & {\color[HTML]{008F0A} -14.10} & {\color[HTML]{008F0A} -14.96} & {\color[HTML]{008F0A} -7.85} & {\color[HTML]{008F0A} -10.43} & {\color[HTML]{008F0A} -10.32} & {\color[HTML]{008F0A} -21.15} & {\color[HTML]{008F0A} -18.30} & {\color[HTML]{008F0A} -16.22} & {\color[HTML]{008F0A} -20.21} & {\color[HTML]{008F0A} -18.24} & {\color[HTML]{008F0A} -16.07} & {\color[HTML]{008F0A} -16.73} & {\color[HTML]{008F0A} -15.03} & \multicolumn{1}{c|}{{\color[HTML]{008F0A} -16.16}} & {\color[HTML]{008F0A} -16.86} & {\color[HTML]{008F0A} -16.86} & {\color[HTML]{008F0A} -17.37} \\
5. Improved 3 $\rightarrow$ 1 & - & +0.59 & {\color[HTML]{008F0A} -0.72} & {\color[HTML]{008F0A} -0.88} & {\color[HTML]{FF0000} +3.03} & {\color[HTML]{008F0A} -0.74} & {\color[HTML]{008F0A} -0.52} & {\color[HTML]{FF0000} +1.98} & {\color[HTML]{FF0000} +2.73} & {\color[HTML]{FF0000} +2.64} & {\color[HTML]{FF0000} +0.98} & {\color[HTML]{FF0000} +1.61} & {\color[HTML]{FF0000} +0.88} & {\color[HTML]{FF0000} +2.01} & {\color[HTML]{FF0000} +4.64} & \multicolumn{1}{c|}{{\color[HTML]{FF0000} +3.95}} & {\color[HTML]{FF0000} +1.51} & {\color[HTML]{FF0000} +1.12} & {\color[HTML]{FF0000} +0.97} \\ \hline
1. PV-RCNN\cite{pvrcnn} & Full & 92.10 & 84.36 & 82.48 & 93.02 & 90.32 & 88.53 & 63.12 & 54.84 & 51.78 & 65.18 & 59.41 & 54.51 & 89.10 & 70.38 & 66.01 & 93.45 & 74.53 & 70.10 \\
2. PV-RCNN\cite{pvrcnn} & Sparse (20\%) & 74.25 & 64.72 & 62.78 & 82.19 & 77.23 & 76.25 & 54.97 & 49.89 & 45.62 & 58.45 & 53.37 & 49.94 & 81.73 & 61.27 & \multicolumn{1}{c|}{56.90} & 84.49 & 65.65 & 61.24 \\
3. SS3D++ (Ours) & Sparse (20\%) & 92.54 & 84.40 & 82.28 & 95.91 & 90.59 & 88.43 & 58.94 & 54.44 & 49.66 & 61.76 & 57.55 & 52.89 & 91.61 & 76.21 & \multicolumn{1}{c|}{71.46} & 94.62 & 78.18 & 73.62 \\
4. Improved 2 $\rightarrow$ 1 & - & {\color[HTML]{008F0A} -17.85} & {\color[HTML]{008F0A} -19.64} & {\color[HTML]{008F0A} -19.70} & {\color[HTML]{008F0A} -10.83} & {\color[HTML]{008F0A} -13.09} & {\color[HTML]{008F0A} -12.28} & {\color[HTML]{008F0A} -8.15} & {\color[HTML]{008F0A} -4.95} & {\color[HTML]{008F0A} -6.16} & {\color[HTML]{008F0A} -6.73} & {\color[HTML]{008F0A} -6.04} & {\color[HTML]{008F0A} -4.57} & {\color[HTML]{008F0A} -7.37} & {\color[HTML]{008F0A} -9.11} & \multicolumn{1}{c|}{{\color[HTML]{008F0A} -9.11}} & {\color[HTML]{008F0A} -8.96} & {\color[HTML]{008F0A} -8.88} & {\color[HTML]{008F0A} -8.86} \\
5. Improved 3 $\rightarrow$ 1 & - & {\color[HTML]{FF0000} +0.44} & {\color[HTML]{FF0000} +0.04} & {\color[HTML]{008F0A} -0.20} & {\color[HTML]{FF0000} +2.89} & {\color[HTML]{FF0000} +0.27} & {\color[HTML]{008F0A} -0.10} & {\color[HTML]{008F0A} -4.18} & {\color[HTML]{008F0A} -0.40} & {\color[HTML]{008F0A} -2.12} & {\color[HTML]{008F0A} -3.42} & {\color[HTML]{008F0A} -1.86} & {\color[HTML]{008F0A} -1.62} & {\color[HTML]{FF0000} +2.51} & {\color[HTML]{FF0000} +5.83} & \multicolumn{1}{c|}{{\color[HTML]{FF0000} +5.45}} & {\color[HTML]{FF0000} +1.17} & {\color[HTML]{FF0000} +3.65} & {\color[HTML]{FF0000} +3.52} \\ \hline
1. Voxel-RCNN\cite{voxelrcnn} & Full & 92.38 & 85.29 & 82.86 & 95.52 & 91.25 & 88.99 & - & - & - & - & - & - & - & - & \multicolumn{1}{c|}{-} & - & - & - \\
2. Voxel-RCNN\cite{voxelrcnn} & Sparse (20\%) & 75.55 & 64.67 & 62.43 & 83.34 & 76.75 & 73.50 & - & - & - & - & - & \multicolumn{1}{c|}{-} & - & - & \multicolumn{1}{c|}{-} & - & - & - \\
3. SS3D++ (Ours) & Sparse (20\%) & 93.59 & 83.83 & 82.79 & 96.71 & 91.70 & 89.26 & - & - & - & - & - & \multicolumn{1}{c|}{-} & - & - & \multicolumn{1}{c|}{-} & - & - & - \\
4. Improved 2 $\rightarrow$ 1 & - & {\color[HTML]{008F0A} -16.83} & {\color[HTML]{008F0A} -20.62} & {\color[HTML]{008F0A} -20.43} & {\color[HTML]{008F0A} -12.18} & {\color[HTML]{008F0A} -14.50} & {\color[HTML]{008F0A} -15.49} & - & - & - & - & - & \multicolumn{1}{c|}{-} & - & - & \multicolumn{1}{c|}{-} & - & - & - \\
5. Improved 3 $\rightarrow$ 1 & - & {\color[HTML]{FF0000} +1.21} & {\color[HTML]{008F0A} -1.46} & {\color[HTML]{008F0A} -0.07} & {\color[HTML]{FF0000} +1.19} & {\color[HTML]{FF0000} +0.45} & {\color[HTML]{FF0000} +0.27} & - & - & - & - & - & \multicolumn{1}{c|}{-} & - & - & \multicolumn{1}{c|}{-} & - & - & - \\ \hline
\end{tabular}}
\label{tab:tab1}
\end{table*}

\section{Experiments}
\label{experiment}
\subsection{Datasets and Evaluation Metrics}
Following the SOTA methods \cite{pvrcnn++,agonet}, we evaluate our SS3D++ on two widely acknowledged 3D object detection datasets: KITTI \cite{kitti} and Waymo Open Dataset \cite{waymo}.

\textbf{KITTI Dataset}
The KITTI 3D and BEV object detection benchmark \cite{kitti} are widely used for performance evaluation.
There are 7,481 samples for training and 7,518 samples for testing and we further divide the training samples into a \emph{train} split of 3,712 samples and a \emph{val} split of 3,769 samples as a common practice \cite{pvrcnn}. 
In addition, due to the occlusion and truncation levels of objects, the KITTI benchmark has three difficulty levels in the evaluation: easy, moderate, and hard.
As for sparsely annotated dataset generation, we randomly keep one annotated object in each 3D scene from \emph{train} split to generate the extremely sparse split. 
Compared with the full annotation of all objects on KITTI, the extremely sparse split only needs to be annotated with 20\% of objects.
For fair comparisons with previous methods, we report the mAP with 40 or 11 recall positions ~\cite{weaklysuper,pvrcnn++}, with a 3D overlap threshold of 0.7, 0.5, and 0.5 for the three classes: car, pedestrian, and cyclist, respectively.
All the models are evaluated on the \emph{val} split.

\textbf{Waymo Open Dataset}
The Waymo Open Dateset is a large-scale benchmark that includes point cloud and RGB images from a LiDAR sensor and five cameras, respectively.
The dataset contains 798 training sequences with around 160k point cloud samples, and 202 validation sequences with 40k point cloud samples, which are annotated in a 360-degree field.
Based on an IoU threshold of 0.7 for vehicles and 0.5 for pedestrians/cyclists. The official 3D detection evaluation metrics including 3D bounding box mean average precision (mAP) and mAP weighted by heading accuracy (mAPH) are used to benchmark the performance for objects of two difficulty levels (LEVEL\_1 and LEVEL\_2). 
Following the experiments setting of PV-RCNN++\cite{pvrcnn++} in OpenPCDet \cite{openpc}, we sample 20\% of data (about 32k samples) with a frame interval of 5, which includes 1,412k annotated 3D bounding boxes.
Following sparsely annotated dataset generation in the KITTI dataset, we randomly keep one annotated object in each scene to generate the extremely sparse split.
Compared with the full annotation of all objects on the Waymo dataset, the extremely sparse split only needs to be annotated with about 2.2\% of objects.

\subsection{Implementation Details}
We implement our method following OpenPCDet \cite{openpc} with the extremely sparse split, and keep the default supervised loss and configurations as the used detector. 
At the training stage, for the KITTI and Waymo, we also adopt the default optimization setting of the original detector (e.g., ADAM optimizer and cosine annealing learning rate\cite{cos}) with the total rounds of 10 and 6, respectively. 
We set the low score threshold $\tau_l$ as 0.01 in reliable background selection.
In our global augmentation, we randomly flip each scene along X-axis and Y-axis with 0.5 probability, and then scale it with a uniformly sampled factor from $[0.8,1.2]$.
Finally, we rotate the point cloud around Z-axis with a random angle sampled from $\left[-\frac{\pi}{4}, \frac{\pi}{4}\right]$.

\begin{table*}[t]
\centering
\caption{Comparison with SS3D \cite{liu2022ss3d} trained with extremely sparse split on KITTI \emph{val} split. The 3D detection and BEV detection are evaluated by mean average precision with 11 recall positions. The IoU thresholds are 0.7 for car and 0.5 for cyclist, respectively.}	
\resizebox{1.0\linewidth}{!}{%
\begin{tabular}{c|c||ccc|ccc|ccc|ccc|c}
\hline
\multirow{2}{*}{Method} & \multirow{2}{*}{Data} & \multicolumn{3}{c|}{3D Detection (Car)} & \multicolumn{3}{c|}{BEV Detection (Car)} & \multicolumn{3}{c|}{3D Detection (Cyc.)} & \multicolumn{3}{c|}{BEV Detection (Cyc.)} & \multirow{2}{*}{Avg.} \\ 
 &  & Easy & Mod & Hard & Easy & Mod & Hard & Easy & Mod & Hard & Easy & Mod & Hard &  \\ \hline
SS3D (PointRCNN-based) & Sparse (20\%) & 87.18 & 77.10 & 76.13 & 89.74 & 87.41 & 85.71 & 86.62 & 73.22 & 66.92 & 87.21 & 74.27 & 71.54 & 80.25 \\
SS3D++ (PointRCNN-based) & Sparse (20\%) & \textbf{88.63} & \textbf{77.90} & \textbf{77.13} & \textbf{90.26} & \textbf{87.65} & \textbf{86.10} & \textbf{86.73} & \textbf{74.81} & \textbf{70.28} & \textbf{87.76} & \textbf{74.64} & \textbf{71.81} & \textbf{81.14} \\
\emph{Improvements} & - & +1.45 & +0.80 & +1.00 & +0.52 & +0.24 & +0.39 & +0.11 & +1.59 & +3.36 & +0.55 & +0.37 & +0.27 & {+0.89} \\ \hline
SS3D (PV-RCNN-based) & Sparse (20\%) & \textbf{89.49} & 79.30 & 78.28 & \textbf{90.45} & 87.98 & 87.00 & 88.01 & 70.35 & 67.40 & 89.72 & 72.33 & 70.14 & 80.87 \\
SS3D++ (PV-RCNN-based) & Sparse (20\%) & 89.32 & \textbf{83.42} & \textbf{78.67} & 90.23 & \textbf{87.91} & \textbf{87.33} & \textbf{88.33} & \textbf{76.13} & \textbf{70.72} & \textbf{93.99} & \textbf{77.54} & \textbf{72.42} & \textbf{82.89} \\
\emph{Improvements} & - & -0.17 & +4.12 & +0.39 & -0.22 & -0.07 & +0.33 & +0.32 & +5.78 & +3.32 & +4.27 & +5.21 & +2.28 & +2.02
\\ \hline
\end{tabular}%
}
\label{tab:tab1_comparison_original}
\end{table*}

\subsection{Comparisons with Fully-Supervised Methods}

\begin{figure}[h]
    \centering
    \includegraphics[width=\linewidth]{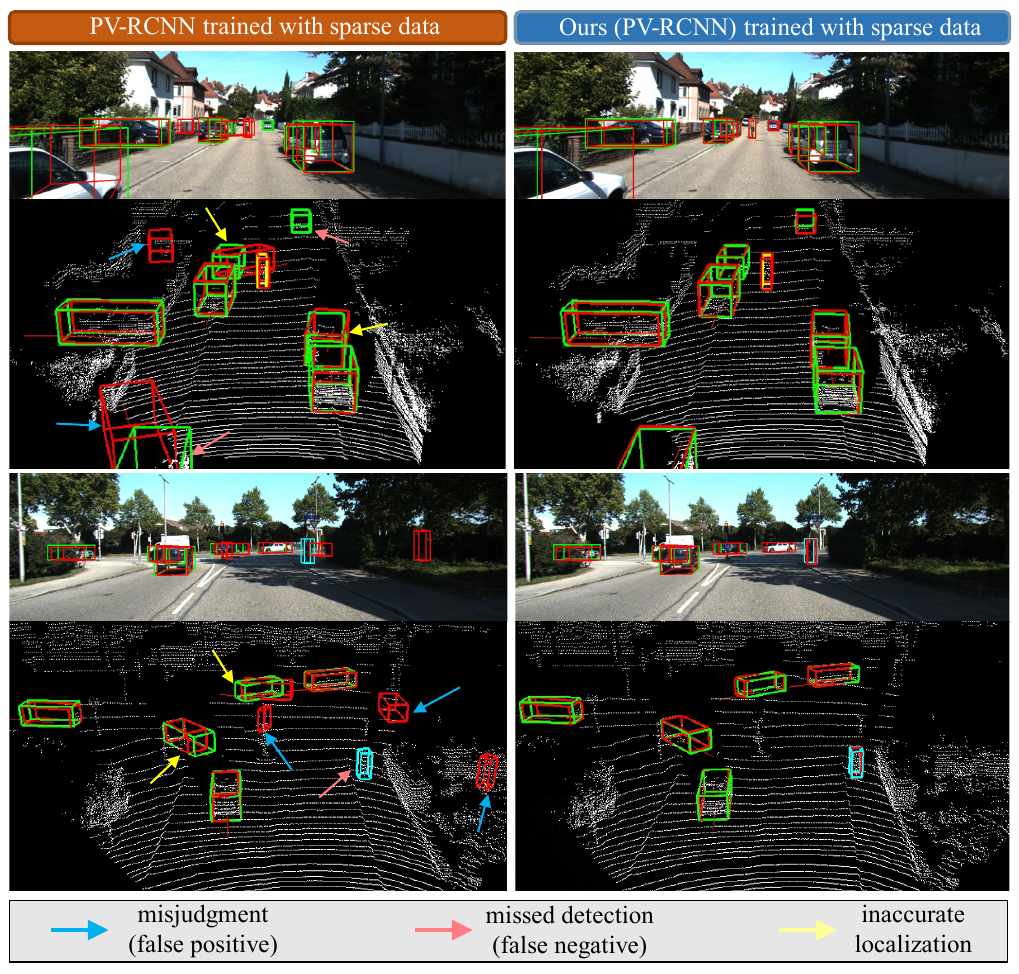}
    \caption{Qualitative comparison results of the PV-RCNN~\cite{pvrcnn} and our SS3D++ method (PV-RCNN-based) trained with sparsely-annotated data. 
    The ground truth 3D bounding boxes of cars, cyclists, and pedestrians are drawn in green, yellow, and cyan, respectively. 
    We set the predicted bounding boxes in red and project boxes in the point cloud back onto the color images for visualization.
    }
    \label{fig:vis} 
\end{figure}

\textbf{Comparison with fully-supervised methods on KITTI.}
To evaluate our SS3D++ framework on the highly-competitive KITTI dataset, we compare the proposed method with four SOTA fully-supervised methods: PointRCNN \cite{pointrcnn}, Part-A$^2$ \cite{parta2}, PV-RCNN \cite{pvrcnn}, Voxel-RCNN \cite{voxelrcnn}, with fully-annotated \emph{train} split and the extremely sparse \emph{train} split, respectively, where these detectors trained on the extremely sparse split are used as the initial detectors of our method.
The results of different methods are shown in Tab. \ref{tab:tab1}.
From this table, it can be seen that the four fully-supervised detectors trained on the sparsely annotated dataset produce a significant performance degradation due to the extremely inexact and incomplete supervision. For example, PointRCNN has an average performance decrease of approximately 25\% points on ``car'' 3D detection, 21\%  points on ``pedestrian'' 3D detection, and 25\%  points on ``cyclist'' 3D detection. Even for the competitive PV-RCNN detector, it exhibits an average decline of about 19\% points on ``car'' 3D detection, 6\%  points on ``pedestrian'' 3D detection, and 8\%  points on ``cyclist'' 3D detection. In the left column of Fig. \ref{fig:vis}, we visualize the detection results of PV-RCNN on the KITTI dataset. As shown, there exist some typical mistakes for the PV-RCNN detector, e.g., the background wrongly detected as an object (indicated by blue arrows), missing object detection (pink arrows), inaccurate localization (yellow arrows), etc.

\begin{table*}[]
\caption{Comparison on 202 validation sequences with different detectors trained with fully annotated and extremely sparse annotated Waymo Open Dataset. 
All detectors are implemented based on \cite{openpc}. \dag: with centerhead \cite{centerpoint}.
}
\resizebox{\textwidth}{!}{
\begin{tabular}{c|c||cc|cc|cc|cc|cc|cc|c}
\hline
\multirow{2}{*}{Method} & \multirow{2}{*}{Data} & \multicolumn{2}{c|}{Veh.(LEVEL\_1)} & \multicolumn{2}{c|}{Veh.(LEVEL\_2)} & \multicolumn{2}{c|}{Ped.(LEVEL\_1)} & \multicolumn{2}{c|}{Ped.(LEVEL\_2)} & \multicolumn{2}{c|}{Cyc.(LEVEL\_1)} & \multicolumn{2}{c|}{Cyc.(LEVEL\_2)} & \multirow{2}{*}{Avg.} \\
 &  & mAP & mAPH & mAP & mAPH & mAP & mAPH & mAP & mAPH & mAP & mAPH & mAP & mAPH &  \\ \hline
PartA$^2$\cite{parta2} & Full & 74.66 & 74.12 & 65.82 & 65.32 & 71.71 & 62.24 & 62.46 & 54.06 & 66.53 & 65.18 & 64.05 & 62.75 & 65.74 \\
PartA$^2$\cite{parta2} & Sparse {(2.2\%)}  & 43.95 & 42.40 & 38.35 & 37.00 & 24.27 & 19.51 & 20.37 & 16.36 & 49.92 & 48.46 & 48.02 & 46.61 & 36.27 \\
Ours (PartA$^2$-based) & Sparse {(2.2\%)}  & 66.27 & 65.35 & 58.03 & 57.02 & 52.54 & 45.68 & 44.37 & 38.46 & 60.71 & 58.80 & 58.12 & 56.67 & 55.17 \\ \hline
$^\dag$Voxel-RCNN \cite{voxelrcnn} & Full & 76.13 & 75.66 & 68.18 & 67.74 & 78.20 & 71.98 & 69.29 & 63.59 & 70.75 & 69.68 & 68.25 & 67.21 & 70.56 \\
$^\dag$Voxel-RCNN \cite{voxelrcnn} & Sparse {(2.2\%)}  & 47.83 & 46.32 & 41.83 & 40.51 & 25.88 & 20.73 & 21.78 & 17.44 & 50.29 & 48.48 & 48.38 & 46.65 & 38.01 \\
Ours(Voxel-RCNN-based) & Sparse {(2.2\%)}  & 65.99 & 65.29 & 57.50 & 56.90 & 53.44 & 48.89 & 46.41 & 42.37 & 60.13 & 58.93 & 57.40 & 56.26 & 55.79 \\ \hline
PV-RCNN++ \cite{pvrcnn++} & Full & 77.82 & 77.32 & 69.07 & 68.62 & 77.99 & 71.36 & 69.92 & 63.74 & 71.80 & 70.71 & 69.31 & 68.26 & 71.33 \\
PV-RCNN++ \cite{pvrcnn++} & Sparse {(2.2\%)}  & 29.96 & 29.20 & 25.95 & 25.29 & 12.26 & 10.35 & 10.20 & 8.61 & 43.66 & 42.54 & 41.99 & 40.91 & 26.74 \\
Ours (PV-RCNN++-based) & Sparse {(2.2\%)}  & 65.73 & 64.96 & 57.34 & 56.67 & 53.62 & 48.51 & 45.33 & 41.01 & 59.78 & 58.36 & 57.06 & 55.72 & 55.34 \\ \hline
CenterPoint \cite{centerpoint} & Full & 71.33 & 70.76 & 63.16 & 62.65 & 72.09 & 65.49 & 64.27 & 58.23 & 68.68 & 67.39 & 66.11 & 64.87 & 66.25 \\
CenterPoint \cite{centerpoint} & Sparse {(2.2\%)}  & 32.98 & 32.33 & 28.82 & 28.24 & 20.22 & 16.91 & 17.30 & 14.47 & 28.84 & 27.90 & 27.74 & 26.84 & 25.22 \\
Ours(CenterPoint-based) & Sparse {(2.2\%)} & 63.49 & 62.93 & 55.30 & 54.81 & 48.93 & 43.89 & 41.53 & 37.22 & 51.87 & 50.81 & 50.07 & 49.04 & 50.82 \\ \hline
\end{tabular}}
\label{tab:tab1waymo}
\end{table*}

Our SS3D++ method is detector-agnostic and can be easily used to further improve these off-the-shelf 3D fully-supervised detectors with sparse supervision. Tab. \ref{tab:tab1} shows that our SS3D++ method can substantially boost the detection performance of different fully-supervised detectors for each category across different difficulty levels, and approaches or even surpasses the results of these methods with full supervision. This highlights the significant advantage of the proposed method in eliminating the negative influence of sparse supervision.
The visual qualitative analysis of detection results is shown in the right column of Fig. \ref{fig:vis}. As can be seen, the SS3D++ method effectively overcomes the mistakes of the PV-RCNN detector, and achieves fine detection results.
The visualization results of other detectors can be found in Appendix Fig. \ref{fig:app_kitti_vis}. 
This further validates the effectiveness of iterations between detector amelioration and confident fully-annotated scene generation.

Tab. \ref{tab:tab1_comparison_original} further compares the SS3D++ method with our previous conference method SS3D. The novel SS3D++ method outperforms the previous SS3D method by a margin on three different difficulty levels for all categories. These consistent improvements demonstrate the effectiveness of the mutual benefit of our SS3D++ between 3D detector training and confident fully-annotated scene generation.

\begin{table*}[h]
\centering
\caption{Comparison with other sparsely-supervised methods trained with extremely sparse annotated KITTI dataset. The 3D object detection benchmark is evaluated by mean average precision with R40, under IoU thresholds 0.7.}
\resizebox{0.8\linewidth}{!}{
\begin{tabular}{c|c|c||ccc|ccc|c}
\hline
\multirow{2}{*}{Method} & \multirow{2}{*}{Venue} & \multirow{2}{*}{Data} & \multicolumn{3}{c|}{Car - 3D Detection} & \multicolumn{3}{c|}{Car - BEV Detection} & \multirow{2}{*}{Avg.} \\
 & & & Easy & Mod & Hard & Easy & Mod & Hard &  \\ \hline
Voxel-RCNN \cite{voxelrcnn} & AAAI'21 & Sparse (2\%) & 72.5 & 54.9 & 44.8 & 83.6 & 71.4 & 57.7 & 64.15 \\ \hline
Coln \cite{xia2023coin} & ICCV'23 & Sparse (2\%) &  84.5 & 68.4 & 58.0 &  92.3 & 81.0 & 70.2 & 75.73 \\
Coln++ \cite{xia2023coin} & ICCV'23 & Sparse (2\%) &  92.0 & 79.5 & 71.5 & 96.1 & 88.8 & 82.5 & 85.06 \\
SP3D \cite{zhao2025sp3d} & CVPR'25 & Sparse (2\%) & 91.3 & 74.8 & 63.8 & 95.4 & 85.2 & 74.5 &  80.83 \\
SS3D++ (Ours) & - & Sparse (2\%) & \textbf{93.2} & \textbf{80.7} & \textbf{75.9} & \textbf{96.6} & \textbf{90.0} & \textbf{84.6} & \textbf{86.83} 
\\
\hline
\end{tabular}}
\label{tab:annotation_sparse_kitti}
\end{table*}

\begin{table*}[h]
\centering
\caption{Comparison with Coln trained with extremely sparse annotated Waymo Open Dataset. 
}
\resizebox{\linewidth}{!}{
\begin{tabular}{c|c||cc|cc|cc|cc|cc|cc|c}
\hline
 &  & \multicolumn{2}{c|}{Veh. (LEVEL 1)} & \multicolumn{2}{c|}{Veh. (LEVEL 2)} & \multicolumn{2}{c|}{Ped. (LEVEL 1)} & \multicolumn{2}{c|}{Ped. (LEVEL 2)} & \multicolumn{2}{c|}{Cyc. (LEVEL 1)} & \multicolumn{2}{c|}{Cyc. (LEVEL 2)} &  \\
\multirow{-2}{*}{Methods} & \multirow{-2}{*}{Data} & mAP & mAPH & mAP & mAPH & mAP & mAPH & mAP & mAPH & mAP & mAPH & mAP & mAPH & \multirow{-2}{*}{Avg.} \\ \hline
CenterPoint \cite{centerpoint} & Fully & 71.33 & 70.76 & 63.16 & 62.65 & 72.09 & 65.49 & 64.27 & 58.23 & 68.68 & 67.39 & 66.11 & 64.87 & 66.25 \\ \hline
CenterPoint (From Coln++) & Sparse (2.2\%) & {32.15} & {31.55} & {27.97} & {27.45} & {25.66} & {21.65} & {22.00} & {18.56} & {59.25} & {57.84} & {57.22} & {55.86} & {36.43} \\
Coln++ \cite{xia2023coin} & Sparse (2.2\%) & {48.25} & {47.60} & {41.82} & {41.25} & {28.25} & {24.28} & {23.79} & {20.45} & {63.99} & {62.60} & {61.71} & {60.37} & {43.69} \\
\emph{Improvements} & - & +16.10 & +16.05 & +13.85 & +13.80 & +2.59 & +2.63 & +1.79 & +1.89 & +4.74 & +4.76 & +4.49 & +4.51 & +7.26 \\ \hline
CenterPoint \cite{centerpoint} (Our produced) & Sparse (2.2\%) & 32.98 & 32.33 & 28.82 & 28.24 & 20.22 & 16.91 & 17.30 & 14.47 & 28.84 & 27.90 & 27.74 & 26.84 & 25.22 \\
SS3D++ (Ours) & Sparse (2.2\%) & 63.49 & 62.93 & 55.30 & 54.81 & 48.93 & 43.89 & 41.53 & 37.22 & 51.87 & 50.81 & 50.07 & 49.04 & 50.82 \\
\emph{Improvements} & - & +30.51 & +30.60 & +26.48 & +26.57 & +28.71 & +26.98 & +24.23 & +22.75 & +23.03 & +22.91 & +22.33 & +22.20 & +25.61 \\ \hline
\end{tabular}}
\label{tab:annotation_coln++_waymo}
\end{table*}

\begin{table*}[t]
\centering
\caption{Performance comparison with state-of-the-art semi-supervised methods on the Waymo Open Dataset with 202 validation sequences for the 3D detection. \textbf{Bold} text and \underline{underlined} text represent the best performance and the second-best performance, respectively. The ``Improvement'' indicates the difference between the best and the second best.
}
\label{tab:waymo_result_semi}
\resizebox{\textwidth}{!}{
\begin{tabular}{c|c||cc|cc|cc|cc|cc|cc|c}
\hline
\multirow{2}{*}{\begin{tabular}[c]{@{}c@{}}1\% Annotation \\ Data\end{tabular}} & \multirow{2}{*}{Venue} & \multicolumn{2}{c|}{Veh. (LEVEL 1)} & \multicolumn{2}{c|}{Veh. (LEVEL 2)} & \multicolumn{2}{c|}{Ped. (LEVEL 1)} & \multicolumn{2}{c|}{Ped. (LEVEL 2)} & \multicolumn{2}{c|}{Cyc. (LEVEL 1)} & \multicolumn{2}{c|}{Cyc. (LEVEL 2)} & \multirow{2}{*}{Avg.} \\
 &  & mAP & mAPH & mAP & mAPH & mAP & mAPH & mAP & mAPH & mAP & mAPH & mAP & mAPH &  \\ \hline
PV-RCNN \cite{pvrcnn} & CVPR'20 & 48.5 & 46.2 & 45.5 & 43.3 & 30.1 & 15.7 & 27.3 & 15.9 & 4.5 & 3.0 & 4.3 & 2.9 & 23.93 \\ \hline
DetMatch \cite{park2022detmatch} & ECCV'22 & 52.2 & 51.1 & 48.1 & 47.2 & 39.5 & 18.9 & 35.8 & 17.1 & - & - & - & - & 38.74 \\
HSSDA \cite{liu2023hierarchical} & CVPR'23 & 56.4 & 53.8 & 49.7 & 47.3 & 40.1 & 20.9 & 33.5 & 17.5 & 29.1 & 20.9 & 27.9 & 20.0 & 34.76 \\
MT \cite{mt} & ICCV'25 & 59.8 & 57.7 & 52.0 & 50.2 & 44.0 & 22.7 & 37.5 & 19.3 & 35.2 & 19.5 & 33.9 & 18.7 & 37.54 \\
A-Teacher \cite{ateacher} & CVPR'24 & 56.5 & 54.5 & 49.2 & 47.5 & \underline{48.1} & \underline{27.3} & \underline{40.8} & \underline{23.1} & 35.1 & \underline{27.1} & 33.7 & 26.1 & 39.08 \\
PPTM \cite{pptm} & AAAI'24 & \underline{61.5} & \underline{59.8} & \underline{53.7} & \underline{52.2} & 43.1 & 22.3 & 36.3 & 18.8 & \underline{35.7} & 17.9 & \underline{35.7} & \underline{34.3} & \underline{39.28} \\ \hline
SS3D++ (Ours) & - & \textbf{62.1} & \textbf{61.3} & \textbf{54.2} & \textbf{53.5} & \textbf{50.9} & \textbf{45.4} & \textbf{43.0} & \textbf{38.4} & \textbf{52.8} & \textbf{51.5} & \textbf{50.8} & \textbf{49.5} & \textbf{51.12} \\
\emph{Improvements} & - & +0.6 & +1.5 & +0.5 & +1.3 & +2.8 & +18.1 & +2.2 & +15.3 & +17.1 & +24.4 & +15.1 & +15.2 & +9.51 \\ \hline
\end{tabular}}
\end{table*}

\noindent\textbf{Comparison with fully-supervised methods on Waymo.} 
The Waymo dataset is larger than the KITTI dataset and the scene is very crowded, with an average of 44 objects in each scene.
Thus it is relatively difficult and expensive to provide full precise annotations. 
The proposed framework of using extremely sparse annotation can effectively reduce the workload to a greater extent (about 2.2\%  points). However, such sparse annotation significantly deteriorates the existing fully-supervised detectors. As shown in Tab. \ref{tab:tab1waymo}, PartA$^2$ and Voxel-RCNN have an average performance decrease of beyond 29\% and 32\%  points, respectively. To further verify the generalizability of our SS3D++, we also replace the base detector with two latest networks that are different from those in the KITTI experiment, \emph{i.e.,} PV-RCNN++\cite{pvrcnn++} and CenterPoint \cite{centerpoint}. While they encounter an average performance decrease of beyond 44\% and 41\%  points, respectively. 

Nonetheless, our SS3D++ method can still achieve about 80\% performance compared with the detectors trained with full supervision and improves the average detection performance of PartA$^2$, Voxel-RCNN, PV-RCNN++, and CenterPoint by 18.90\%, 17.76\%, 28.60\%, 25.60\% points, 
respectively. This demonstrates that SS3D++ method can achieve a consistent performance improvement across different datasets/detectors. Observing that we only have access to only 2.2\% precise annotations, this implies that the SS3D++ method is hopeful to be readily applied to real-world 3D object detection that requires low-cost annotations. The qualitative comparison results on the Waymo Open dataset can be found in Appendix Fig. \ref{fig:app_kitti_vis}.

\begin{figure}[h]
    \centering
    \includegraphics[width=\linewidth]{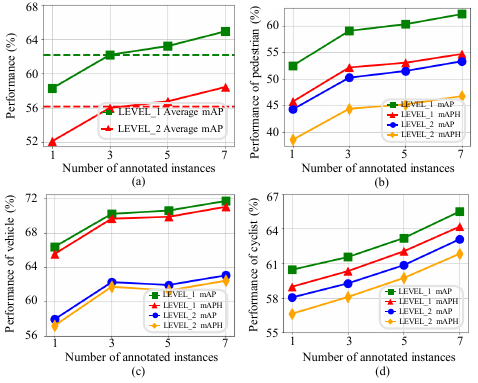}
    \caption{Detection performance against the number of annotated instances per scene on the Waymo dataset.
    }
    \label{fig:waymo_ratio_data} 
\end{figure}

Due to the crowded nature of the Waymo dataset, the distribution of objects within the scenes is usually dense. When we only annotate one instance per scene, it poses significant challenges for detection learning. The large number of missing annotated instances results in seriously incomplete supervision, which causes a detrimental performance. We further attempt to annotate more instances for each scene to improve the performance. Fig. \ref{fig:waymo_ratio_data} illustrates the varying curve of average detection performance as the number of annotated instances increase per scene. It can be observed that we could achieve about 90\% of the performance achieved by fully supervised methods (as depicted by the dashed line in Fig. \ref{fig:waymo_ratio_data} (a)) with only additionally annotating two objects per scene. Especially, we only require about $15\times$ less annotation cost compared with full annotation. We also show the varying curve of detection performance in terms of  the vehicle, pedestrian, and cyclist categories in Fig. \ref{fig:waymo_ratio_data} (b, c, d). It can be seen that the detection performance of different categories can be progressively improved as the number of annotated instances increase per scene. This implies that the SS3D++ method is capable of boosting performance by employing more annotation information.

\subsection{Comparisons with SOTA Methods} \label{weakly}
\textbf{Comparison with the sparsely-supervised method.}
We have compared our method with recently SOTA methods \cite{xia2023coin,zhao2025sp3d,zhu2025learning} under the sparsely-supervised setting. Tab. \ref{tab:annotation_sparse_kitti} presents the performance comparison on the KITTI dataset. To ensure a fair comparison, we used Voxel-RCNN as the backbone detector with 2\% labeled data and reported the Recall@40 performance at an IoU threshold of 0.7. It is evident that our method outperforms Coln++ across different difficulty levels. Moreover, we achieved the highest average performance across all methods.
Additionally, Tab. \ref{tab:annotation_coln++_waymo} provides the performance comparison on the Waymo dataset~\cite{waymo} under the same sparse annotation settings. Similarly, it can be observed that our SS3D++ significantly outperforms Coln++ in this setting as well, highlighting its robustness and generalization capability across datasets.

\noindent\textbf{Comparison with the semi-supervised method.}
First, we compare our results on the large-scale Waymo dataset \cite{waymo} with current state-of-the-art semi-supervised methods. For fair comparison, we keep all methods trained with the same number of annotated objects and our SS3D++ adopts PV-RCNN as the backbone detector. Remarkably, with just 1\% of the annotation effort, we are able to achieve competitive results, as illustrated in Tab. \ref{tab:waymo_result_semi}. As can be seen from the last row, our method outperforms the semi-supervised methods with best performance across all categories, with an average improvement of 9.5 points.

\begin{figure}[t]
    \centering
    \includegraphics[width=\linewidth]{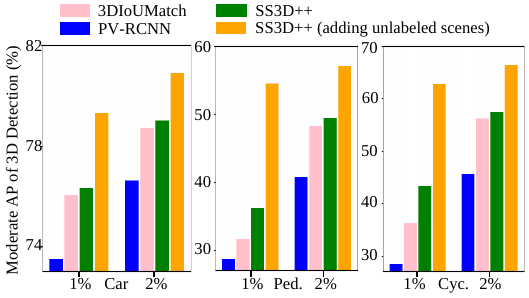}
    \caption{
    Performance comparison on KITTI \emph{val} split with semi-supervised 3DIoUMatch \cite{3dioumatch} method trained with 1\% or 2\% labeled data. Both SS3D++ and 3DIoUMatch methods are based on PV-RCNN \cite{pvrcnn}. We report the mAP with 40 recall positions for a fair comparison.
    }
    \label{fig:com_semi} 
\end{figure}

\begin{table*}[t]
\centering
\caption{Performance comparison on KITTI \emph{val} split with center-click weakly-supervised WS3D \cite{weaklysuper} method. We report the mAP with 11 recall positions.
`*' denotes the scenes with center-click annotations and `$\dag$' denotes precisely-annotated instances. 
}
\resizebox{14.5cm}{!}{
\begin{tabular}{c|c||ccc|ccc|l}
\hline
\multirow{2}{*}{Training Data} & \multirow{2}{*}{Method} & \multicolumn{3}{c|}{3D Detection (Car)} & \multicolumn{3}{c|}{BEV Detection (Car)} & \multirow{2}{*}{Avg.} \\ 
 &  & Easy & Mod & Hard & Easy & Mod & Hard &  \\ \hline
500* scenes + 534$^\dag$ instances & WS3D \cite{weaklysuper} & 85.04 & 75.94 & 74.38 & 88.95 & 85.83 & 85.03 & 82.52 \\ \hline
\multirow{4}{*}{534$^\dag$ instances} & Ours (PointRCNN-based \cite{pointrcnn}) & 84.69 & 71.93 & 67.06 & 89.57 & 84.07 & 78.81 & 79.36 \\
 & Ours (PartA$^2$-based \cite{parta2}) & 88.56 & 77.91 & 72.46 & 90.00 & 86.45 & 82.28 & 82.94 \\
 & Ours (Voxel-RCNN-based \cite{voxelrcnn}) & 88.37 & 77.72 & 75.07 & 90.19 & 87.33 & 84.28 & 83.83 \\
 & Ours (PV-RCNN-based \cite{pvrcnn}) & 89.26 & 78.8 & 76.55 & 90.19 & 87.24 & 85.05 & 84.52 \\ \hline
\end{tabular}}
\label{tab:tab_weak}
\end{table*}

\begin{figure}[t]
    \centering
    \includegraphics[width=\linewidth]{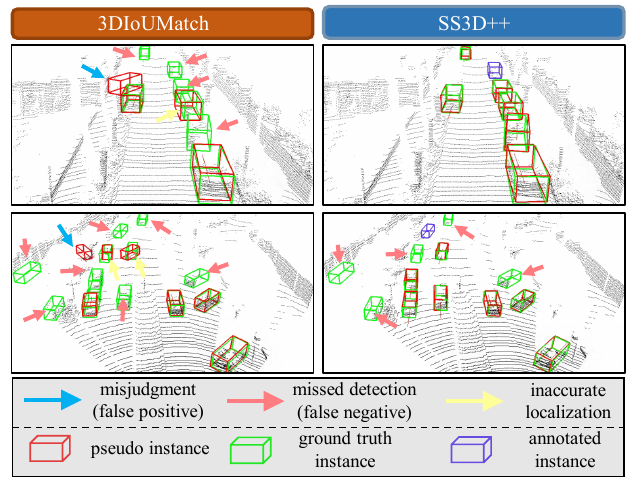}
    \caption{Qualitative comparison of mining pseudo instances by 3DIoUMatch~\cite{3dioumatch} and SS3D++ methods.
    }
    \label{fig:intra_cross_scene} 
\end{figure}

Then, We compare the proposed SS3D+ method with semi-supervised method 3DIoUMatch \cite{3dioumatch} on the KITTI dataset \cite{kitti}.
To make a fair comparison, we also adopt the PV-RCNN as the base detector and keep all methods trained with the same number of annotated objects. 
Specifically, in the KITTI \emph{train} split, there are 3,712 scenes and these scenes contain a total of 17,289 objects for cars, pedestrians, and cyclists. 
For semi-supervised methods, 1\% labeled data means 37 ($3712\times1\%$) scenes, which include an average of 172 ($17289\times1\%$) labeled objects used for training.
So as for 1\% labeled data in our extremely sparse split, we randomly select 172 scenes including 172 labeled objects for training.
We also test the case of 2\% labeled training data for both methods. Generally, our sparse annotation strategy may provide an easier way to annotate objects from point cloud scenes compared to a dense annotation strategy for semi-supervised methods.
The detection results with different ratios of labeled data can be seen in Fig. \ref{fig:com_semi}.
As we can see, our SS3D++ (the green bar) significantly outperforms the current SOTA semi-supervised method, 3DIoUMatch, for three different classes with all ratios of labeled training data split. Fig. \ref{fig:intra_cross_scene} illustrates the comparison of mining pseudo instances by 3DIoUMatch and SS3D++ methods.
The 3DIoUMatch method may easily fall into a suboptimal solution due to limited information transfer from labeled to large discrepancy unlabeled scenes, while the SS3D++ method can encourage easier intra-scene information transfer due to the proposed sparse annotation strategy, in which each scene has one object annotation.
In fact, the results of the green bar in Fig. \ref{fig:com_semi} only use 172 (or 344) scenes for training, while 3DIoUMatch has access to all 3712 scenes involved in training. This illustrates the potential of the SS3D++, and we can further boost the performance of the SS3D++ by utilizing the remaining unlabeled scenes, and obtain more improvements as depicted by the yellow bar in Fig. \ref{fig:com_semi}, outperforming 3DIoUMatch method by a large margin (more details can refer to Sec. \ref{unlabel}).

\noindent\textbf{Comparison with the center-click weakly-supervised method.}
For the weakly-supervised method, WS3D \cite{weaklysuper}, trains the detector with the first 500 scenes with center-click labels, mixed with randomly selected 534 precisely-annotated instances.
Since the standard detectors are not applicable with center-click labels, we only use the same number of (534) precisely-annotated instances randomly chosen from the first 534 scenes to train our proposed SS3D++ method.
From Tab. \ref{tab:tab_weak}, we can easily observe that the SS3D++ method is able to achieve better performance than WS3D for three different base detectors. WS3D may suffer from the significant disparity between point-level weak annotations and 3D precise annotations, which we only have access to precisely-annotated instances, without point-level annotation efforts, which prevents detectors from overfitting the disparity.
Moreover, it is worth noting that we can obtain better performance gains when we use more powerful detectors, \emph{e.g.,} from PointRCNN to PartA$^2$. This highlights the advantage of our detector-agnostic framework, hinting at the potential of our method which employs advanced SOTA detectors to achieve competitive performance.
Besides, the SS3D++ method can employ additional unlabeled scenes to further boost the performance (refer to Sec. \ref{unlabel}), while WS3D cannot. This makes our SS3D++ easily apply to real-world 3D detection.

\subsection{Ablation Study and Analysis} \label{ablation}
In this section, we present a series of ablation studies to analyze the effects of different modules in our proposed SS3D++ method. 
Following the aforementioned setting, all models are trained on the KITTI dataset with an extremely sparse split and evaluated on the \emph{val} split. 
As for the ablation studies on the Waymo dataset, please refer to the Appendix A.
We take PointRCNN\cite{pointrcnn} as an example base detector to conduct our ablation study, and our methods can obtain similar results for other detectors.
Tab. \ref{tab:overall_abl} summarizes the ablation results on our reliable background mining module (RBMM) and confident missing-annotated instances mining module (CMIM).
All results are reported by the mAP with 40 recall positions under moderate difficulty.

\noindent\textbf{Effect of the reliable background mining module.}
In the $1^{st}$ row of Tab. \ref{tab:overall_abl}, we discarded both modules, so it represents the standard PointRCNN detector trained with the extremely sparse split, whose performances have significant degradation compared with full annotations. 
In the $2^{nd}$ row, we leverage the reliable background mining module to extract reliable backgrounds. It can be observed that the performance of the three categories is significantly improved, especially for the ''car'' category, which has shown a substantial increase. 
This is due to that the autonomous driving dataset usually contains the largest number of cars. Therefore, under the sparsely-supervised setting, the detector trained with massive missing-annotated instances like cars may severely degenerate detection performance.
Fortunately, the proposed reliable background mining module could reduce the risk of treating the missing-annotated instances as background, and thereby facilitate the robust training of the detector by eliminating the negative influence of noise background in the point cloud scenes.

\begin{figure*}[t]
    \centering
    \includegraphics[width=\linewidth]{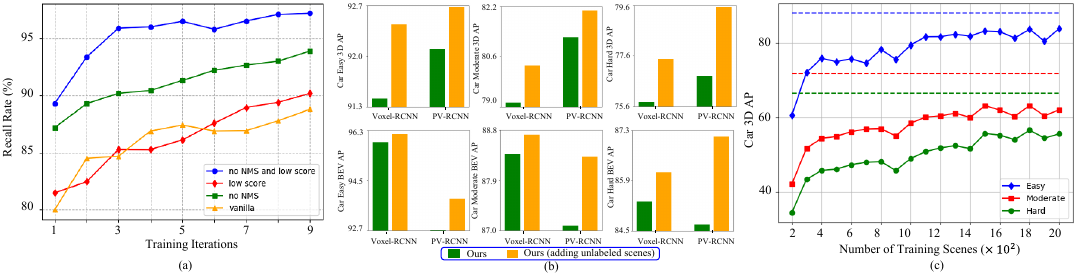}
    \caption{(a): Analysis of removal recall rate for missing-annotated foreground points with different RBMM settings. (b) Whether to add  unlabeled data for performance (R40) comparison when conducting experiments with 534 sparsely annotated scenes. (c) Changing tendencies in terms of AP for SS3D++ with streaming unlabeled scenes (The dash lines represent the results learned from all unlabeled scenes in an offline manner).}
    \label{fig:three_fig}  
\end{figure*}

Furthermore, we analyze different operations of the reliable background mining module via the recall rate of removing noisy foreground points, which indicates how many point clouds belonging to missing-annotated instances in the scene are identified and then removed. The results, as illustrated in Fig. \ref{fig:three_fig} (a), reveal that the joint operations of utilizing a ''low threshold`` and ''No NMS`` bring a substantial improvement in terms of recall rate. In the end, it achieves a 98\% elimination of noisy foreground points and thus reduces the negative impact on the detector caused by massive missing-annotated instances.

\begin{table}[t]
\centering
\caption{Ablation study of different components for our proposed SS3D++ method. We report the mAP with 40 recall positions. RBMM: reliable background mining module. CMIM: confident missing-annotated instances mining module.
}
\begin{tabular}{cc||ccc|c}
\hline
\multirow{2}{*}{RBMM} & \multirow{2}{*}{CMIM} & \multicolumn{3}{c|}{3D Detection (Mod.)} & \multirow{2}{*}{Avg.} \\ 
 &  & Car & Ped. & Cyc. &  \\ \hline
- & - & 54.91 & 32.41 & 46.95 & 44.76 \\
\checkmark & - & 79.35 & 47.26 & 66.67 & 64.43 \\
- & \checkmark & 69.25 & 49.29 & 67.04 & 61.86 \\
\checkmark & \checkmark & 79.85 & 55.98 & 74.13 & 69.99 \\ \hline
\end{tabular}
\label{tab:overall_abl}
\end{table}

\begin{table}
    \centering
    \caption{Ablation study of different components for CMIM of our proposed SS3D++ method. We report the mAP with 40 recall positions. DLF: dynamic loss-based filtering. DCS: dynamic consistency-guided suppression. DCF: density-aware curriculum filtering.}
    \label{tab:cmim_abl} 
    \resizebox{1.0\linewidth}{!}{%
    \begin{tabular}{c|ccc||ccc|c}
    \hline
    \multirow{2}{*}{\emph{{row}}} & \multirow{2}{*}{DLF} & \multirow{2}{*}{DCS} & \multirow{2}{*}{DCF} & \multicolumn{3}{c|}{3D Detection (Mod.)} & \multirow{2}{*}{Avg.} \\
    & &  &  & Car & Ped. & Cyc. &  \\ \hline
    {1} & - & - & - & 79.35 & 47.26 & 66.67 & 64.43 \\
    {2} & \checkmark & - & - & 79.00 & 44.74 & 67.91 & 63.88 \\
    {3} & - & \checkmark & - & 79.62 & 39.90 & 64.10 & 61.21 \\
    {4} & {-} & {-} & {\checkmark} & {74.83} & {27.9} & {52.94} & {51.89} \\
    {5} & \checkmark & \checkmark & - & 78.94 & 49.49 & 71.36 & 66.66 \\
    {6} & {\checkmark} & {-} & {\checkmark} & {78.39} & {49.91} & {72.25} & {66.85} \\
    {7} & {-} & {\checkmark} & {\checkmark} & {77.13} & {45.39} & {72.55} & {65.02} \\
    {8} & \checkmark & \checkmark & \checkmark & 79.85 & 55.98 & 74.13 & 69.99 \\ \hline
    \end{tabular}
    }
\end{table}

\begin{figure}[t]
    \centering
    \includegraphics[width=\linewidth]{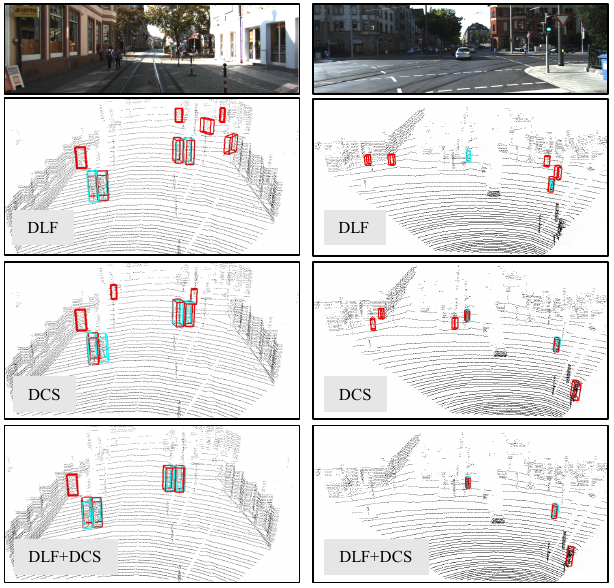}
    \caption{
    Qualitative comparison of mining pseudo instances for different components in the CMIM module. For ease of viewing, we only show the key instances that need to be analyzed. The ground truth 3D bounding boxes and pseudo instances of "pedestrian" are shown in cyan and red, respectively. The combination of DLF and DCS can effectively filter out the most of erroneous pseudo instances.
    }
    \label{fig:abl_vis} 
\end{figure}

\begin{figure}[t]
    \centering
    \includegraphics[width=\linewidth]{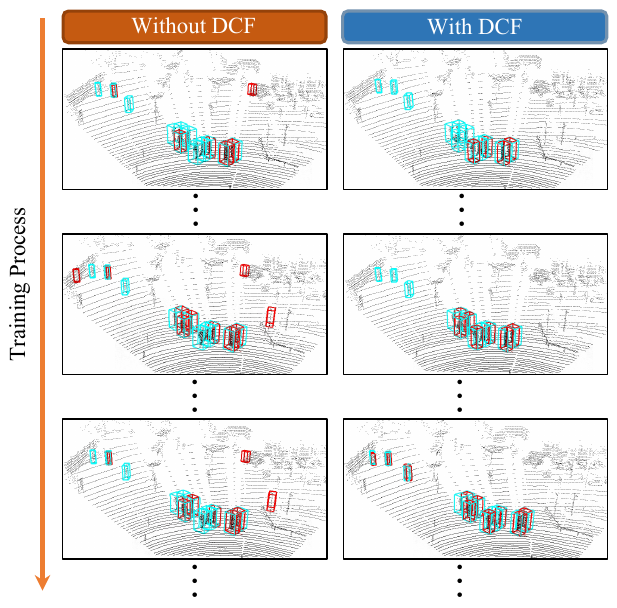}
    \caption{
    Visualization of pseudo instance mining with and without the DCF module. The visualization settings are similar to Fig. \ref{fig:abl_vis}. The design of the DCF enables the discovery of more pseudo instances and significantly reduces the occurrence of false positives.
    }
    \label{fig:dcf_vis}  
\end{figure}

\noindent\textbf{Effect of the confident missing-annotated instance mining module.}
As shown in the $3^{rd}$ row of Tab. \ref{tab:overall_abl}, the confident missing-annotated instance mining module can also improve the baseline by mining high-quality instances. We can observe that the performance of the three categories is significantly improved, especially for the ``pedestrian'' and ``cyclist'' categories, which have shown better improvement compared with only RBMM. Actually, both of ``pedestrian'' and ``cyclist'' categories are relatively hard to detect compared with the ''car'' category, naturally leading to more noisy pseudo-annotated instances. Our multi-criteria sample selection process can effectively filter out noisy pseudo-annotated instances, and thereby brings more robust detection performance. Therefore, it could bring mutual amelioration when we combine CMIM and RBMM ($4^{th}$ row), and achieves better performances than using either of the two.

To comprehend the roles of different components in CMIM, we further conduct a series of ablation experiments, as shown in Tab. \ref{tab:cmim_abl}. 
From the 2$^{nd}$ and 3$^{rd}$ row of Tab. \ref{tab:cmim_abl}, 
we can see that DLF or DCS severely deteriorates the performance of the ``pedestrian'' category. This is because ``pedestrian'' category tends to be smallest object in the training scenes, implying that the pseudo annotations are extremely noisy (even under the full supervision, the mAP score is about 60\%). Both DLF and DCS depend on the quality of pseudo annotations, and thus their performance may degenerate for the ``pedestrian'' category with noisy pseudo annotations. While we combine DLF and DCS strategy ({the 5$^{th}$ row} of Tab. \ref{tab:cmim_abl}), the mAP of the ``pedestrian'' category reaches +2.23\% point compared with the baseline (1$^{st}$ row). As shown in Fig. \ref{fig:abl_vis},  using DLF or DCS alone leads to a significant number of erroneous pseudo instances for "pedestrian". However, combining both components can greatly mitigate this issue.

{It can be observed that using the DCF module alone results in negative effects, as it is challenging to rely solely on density as a criterion for judging incorrect pseudo-labels. However, combining the DCF module with either the DLF or DCS modules leads to performance gains.}
A large performance gain for ``pedestrian'' (+6\% point) can be obtained when we additionally employ the DCF strategy ({from the 5$^{th}$ row to the 8$^{th}$ row} of Tab. \ref{tab:cmim_abl}). Meanwhile, Fig. \ref{fig:dcf_vis} shows the quality improvement of pseudo instances by embedding the DCF with the easy-to-hard curriculum manner.

\subsection{Learning from Unlabeled Training Scenes} \label{unlabel}
We have studied the situation where every training scenes have sparse annotations above. Here, we consider a more practical setting, i.e., we only have access to a small number of training scenes with sparse annotations, and a large pool of unlabeled training scenes. We study two different scenarios exploiting unlabeled training scenes in the following.

\noindent\textbf{Learning from unlabeled training scenes in an offline manner.}
In Sec. \ref{weakly}, our SS3D++ method only employs the sparse annotated scenes. Here, we explore to additionally exploit the remaining unlabeled scenes to train the detector. Observing from the yellow bar in Fig. \ref{fig:com_semi} and Fig. \ref{fig:three_fig} (b), we can see that our SS3D++ method can achieve a large improvement by utilizing additional unlabeled scenes, making it superior over weakly-supervised 3DIoUMatch~\cite{3dioumatch} and WS3D~\cite{weaklysuper} methods. 
Specifically, comparing our method with the 3DIoUMatch \cite{3dioumatch} in terms of the coverage of generated pseudo instances for unlabeled scenes, we find that 3DIoUMatch achieves only 28\% coverage, while our method achieves about 60\% coverage.
The corresponding accuracy of mined instances in terms of the car, pedestrian, and cyclist can reach 92\%, 85\%, and 88\% for the SS3D++ method. This highlights the potential practical value of exploiting unlabeled scenes. Additional analysis of generating pseudo-annotated instances can be found in Appendix Fig. \ref{fig:app_kitti_error_pseudo} and Fig. \ref{fig:app_kitti_interest_pseudo}.
Compared with WS3D~\cite{weaklysuper}, our sparse supervision is easy to annotate and universal, making our method capable of generating confident pseudo annotations for easily obtained unlabeled point cloud scenes.

\noindent\textbf{Learning from unlabeled training scenes in an online manner.}
For an autonomous driving system, it is necessary for the intelligent agent to continually acquire training data and accumulate knowledge, so as to effectively adapt to dynamic environments and driving conditions. Here, we focus on a simple preliminary attempt that incrementally learns the detector from streaming unlabeled data. 
Specifically, we simulate this scenario by considering a system with a memory constraint of 100 point cloud scenes. The initial model is firstly trained on 200 sparsely annotated scenes and subsequently received training scenes in an online manner. We set the online batch size to 100. Our SS3D++ method can generate reliable pseudo-annotated instances for the sequential unlabeled training scenes. After updating the detector, we randomly select the generated full-annotated scenes to update the memory, and remove the remaining training scenes. The detailed algorithm description is summarized in Appendix Algorithm \ref{alg:refine}.

Fig. \ref{fig:three_fig} (c) shows the changing tendencies of mAP in sequentially involved unlabeled training scenes. As can be seen, our method can continually improve the detection performance even when we can only have access to streaming unlabeled scenes with limited memory. This result implies that our method is potentially useful for the autonomous driving system, which is required to perceive the changing environment and incrementally improve itself.

\section{Conclusion}

In this work, we study a novel weakly-annotated framework for 3D object detection from point cloud. Compared with existing densely-annotated information, we only need to annotate one object per scene. This sparse annotation strategy provides an opportunity to substantially reduce the heavy data annotation burden. We design a SS3D++ method that alternately generates high-quality fully-annotated scenes and updates the 3D detector, thereby achieving progressive amelioration of the detection model. We present extensive comparative evaluations of the SS3D++ method on the classic KITTI and more challenging Waymo Open benchmarks, and prove its capability of achieving promising performance with significantly reducing annotation cost. An important characteristic of the proposed method is detector-agnostic, which implies that it can easily benefit from advanced fully-supervised 3D detectors. Besides, the SS3D++ method is able to utilize additional unlabeled training scenes to ameliorate the model. 
The above results show our framework is potentially useful to help make it practical and low-budget for 3D object detection. 

Though impressive results are obtained, there is still large room for further improvement. For example, it is challenging to generate high-quality pseudo-annotated instances for training scenes with crowded objects in our sparsely-supervised framework. Also, designing an effective approach to better make use of massive unlabeled training scenes is an important step towards practice. Especially, given a large number of algorithmic breakthroughs over the past few years, like meta-learning \cite{shu2021learning}, foundation models \cite{bommasani2021opportunities,kirillov2023segment}, etc, we would expect a flurry of innovation towards this promising direction.

\ifCLASSOPTIONcaptionsoff
  \newpage
\fi

{\small
\bibliographystyle{IEEEtran}
\bibliography{egbib}
}

\newpage
\onecolumn

\appendices

\section{Additional experiments on the Waymo dataset} 

We would like to mention that our method is a general framework, producing consistent ablation experiment results across different datasets, such as KITTI \cite{kitti} and Waymo \cite{waymo}. 
We conduct ablation study on the large-scale Waymo dataset, with Voxel-RCNN \cite{voxelrcnn} as our backbone detector, which shows similar results as on KITTI dataset.
Tab. \ref{tab:abl_waymo_1} summarizes the ablation results on our reliable background mining module (RBMM) and confident missing-annotated instances mining module (CMIM).
We report the performance of 3D bounding box mean average precision (mAP) and mAP weighted by heading accuracy (mAPH) for objects of two difficulty levels (LEVEL 1 and LEVEL 2).
In the $1^{st}$ row of Tab. \ref{tab:abl_waymo_1}, we discarded both modules, so it represents the standard fully-supervised detector trained with the extremely sparse split, whose performances have significant degradation compared with full annotations. The proposed reliable background mining module could reduce the risk of treating the missing-annotated instances as background, and thereby facilitate the training under the sparsely-supervised setting.
The confident missing-annotated instance mining module can also improve the baseline by mining high-quality instances, as illustrated in the $3^{rd}$ row of Tab. \ref{tab:abl_waymo_1}.
Using the reliable background mining module alone can lead to some improvements in the pedestrian and cyclist categories, but it results in a decline in vehicle performance. This is because, in the Waymo dataset, vehicles are the most numerous objects within scenes and include categories with significant size variations, such as ``bus'' and ``truck''. This variation makes it challenging to achieve high-quality instances mining with ambiguous background distraction.

\begin{table*}[h]
\centering
\caption{{Ablation study of different components on the Waymo dataset for our proposed SS3D++ method. RBMM: reliable background mining module. CMIM: confident missing-annotated instances mining module.}
}
\resizebox{\textwidth}{!}{
\begin{tabular}{cc||cc|cc|cc|cc|cc|cc|c}
\hline
 &  & \multicolumn{2}{c|}{Veh. (LEVEL 1)} & \multicolumn{2}{c|}{Veh. (LEVEL 2)} & \multicolumn{2}{c|}{Ped. (LEVEL 1)} & \multicolumn{2}{c|}{Ped. (LEVEL 2)} & \multicolumn{2}{c|}{Cyc. (LEVEL 1)} & \multicolumn{2}{c|}{Cyc. (LEVEL 2)} &  \\
\multirow{-2}{*}{RBMM} & \multirow{-2}{*}{CMIM} & mAP & mAPH & mAP & mAPH & mAP & mAPH & mAP & mAPH & mAP & mAPH & mAP & mAPH & \multirow{-2}{*}{Avg.} \\ \hline 
- & - & { 47.83} & { 46.32} & { 41.83} & { 40.51} & { 25.88} & { 20.73} & { 21.78} & { 17.44} & { 50.29} & { 48.48} & { 48.38} & { 46.65} & { 38.01} \\
\checkmark & - & { 63.51} & { 62.92} & { 55.21} & { 54.69} & { 51.32} & { 45.77} & { 43.59} & { 38.83} & { 50.40} & { 49.38} & { 48.47} & { 47.48} & { 50.96} \\
- & \checkmark & { 42.77} & { 42.38} & { 36.81} & { 36.47} & { 33.94} & { 30.97} & { 28.31} & { 25.82} & { 58.73} & { 57.74} & { 56.47} & { 55.52} & { 42.16} \\
\checkmark & \checkmark & { 65.99} & { 65.29} & { 57.50} & { 56.90} & { 53.44} & { 48.89} & { 46.41} & { 42.37} & { 60.13} & { 58.93} & { 57.40} & { 56.26} & { 55.79} \\ \hline
\end{tabular}}
\label{tab:abl_waymo_1}
\end{table*}

Tab. \ref{tab:abl_waymo_2} presents the ablation study results for different components of CMIM within the proposed SS3D++ method, specifically evaluating the impact of Dynamic Loss-based Filtering (DLF), Dynamic Consistency-guided Suppression (DCS), and Density-aware Curriculum Filtering (DCF).
When none of the components (DLF, DCS, DCF) are enabled, it represents the model with reliable background mining module.
When using DLF, DCS, or DCF alone for pseudo instance mining, the average performance declines to varying degrees. This is because the Waymo dataset contains complex and diverse samples, making it difficult to accurately determine pseudo-labels using a single criterion for different classes. Our multi-criteria sample selection process offers varying degrees of improvement, and when all three criteria are applied together, the greatest benefit is achieved.

\begin{table*}[h]
\centering
\caption{{Ablation study of different components for CMIM of our proposed SS3D++ method. DLF: dynamic loss-based filtering. DCS: dynamic consistency-guided suppression. DCF: density-aware curriculum filtering.}
}
\resizebox{\textwidth}{!}{
\begin{tabular}{ccc||cc|cc|cc|cc|cc|cc|c}
\hline
 &  &  & \multicolumn{2}{c|}{Veh. (LEVEL 1)} & \multicolumn{2}{c|}{Veh. (LEVEL 2)} & \multicolumn{2}{c|}{Ped. (LEVEL 1)} & \multicolumn{2}{c|}{Ped. (LEVEL 2)} & \multicolumn{2}{c|}{Cyc. (LEVEL 1)} & \multicolumn{2}{c|}{Cyc. (LEVEL 2)} &  \\
\multirow{-2}{*}{DLF} & \multirow{-2}{*}{DCS} & \multirow{-2}{*}{DCF} & mAP & mAPH & mAP & mAPH & mAP & mAPH & mAP & mAPH & mAP & mAPH & mAP & mAPH & \multirow{-2}{*}{Avg.} \\ \hline
- & - & - & {63.51} & {62.92} & {55.21} & {54.69} & {51.32} & {45.77} & {43.59} & {38.83} & {50.40} & {49.38} & {48.47} & {47.48} & {50.96} \\
\checkmark & - & - & {65.70} & {65.02} & {57.55} & {56.94} & {43.94} & {39.60} & {37.58} & {33.82} & {50.79} & {49.84} & {48.84} & {47.93} & {49.79} \\
- & \checkmark & - & {66.73} & {65.97} & {58.55} & {57.88} & {38.46} & {34.76} & {32.86} & {29.65} & {54.00} & {52.92} & {51.93} & {50.89} & {49.55} \\
- & - & \checkmark & 60.24 & 59.17 & 51.77 & 50.39 & 31.24 & 26.94 & 24.64 & 21.47 & 48.84 & 46.97 & 44.73 & 43.24 & 42.47 \\
\checkmark & \checkmark & - & {65.16} & {64.48} & {57.04} & {56.44} & {51.56} & {46.49} & {44.03} & {39.70} & {59.59} & {58.49} & {57.35} & {56.30} & {54.71} \\
\checkmark & - & \checkmark & 65.78 & 65.17 & 57.49 & 56.39 & 50.35 & 45.92 & 43.73 & 37.66 & 56.48 & 55.12 & 53.98 & 52.79 & 53.40 \\
- & \checkmark & \checkmark & 64.98 & 64.22 & 58.03 & 56.84 & 45.23 & 40.77 & 38.29 & 33.88 & 54.49 & 52.64 & 51.21 & 50.04 & 50.88 \\
\checkmark & \checkmark & \checkmark & {65.99} & {65.29} & {57.50} & {56.90} & {53.44} & {48.89} & {46.41} & {42.37} & {60.13} & {58.93} & {57.40} & {56.26} & {55.79} \\ \hline
\end{tabular}}
\label{tab:abl_waymo_2}
\end{table*}

\subsection{Performance comparison between SS3D and SS3D++}

{
We conduct the experiments on the Waymo dataset to compare SS3D++ with SS3D \cite{liu2022ss3d}. Tab. \ref{tab:waymo_result_ss3d} shows the comparison results, from which it can be seen that the performance of SS3D++ on the Waymo dataset has significantly improved compared to SS3D. SS3D++ shows the highest average performance across all categories and levels, suggesting it is the most robust model among those tested.}

\begin{table*}[h]
\centering
\caption{{Performance comparison with SS3D on the Waymo Open Dataset with 202 validation sequences for the 3D detection. \dag: with centerhead.}
}
\resizebox{\textwidth}{!}{
\begin{tabular}{c|c||cc|cc|cc|cc|cc|cc|c}
\hline
\multirow{2}{*}{Methods} & \multirow{2}{*}{Data} & \multicolumn{2}{c|}{Veh. (LEVEL 1)} & \multicolumn{2}{c|}{Veh. (LEVEL 2)} & \multicolumn{2}{c|}{Ped. (LEVEL 1)} & \multicolumn{2}{c|}{Ped. (LEVEL 2)} & \multicolumn{2}{c|}{Cyc. (LEVEL 1)} & \multicolumn{2}{c|}{Cyc. (LEVEL 2)} & \multirow{2}{*}{Avg.} \\
 &  & mAP & mAPH & mAP & mAPH & mAP & mAPH & mAP & mAPH & mAP & mAPH & mAP & mAPH &  \\ \hline
CenterPoint \cite{centerpoint} & Sparse (2.2\%) & 32.98 & 32.33 & 28.82 & 28.24 & 20.22 & 16.91 & 17.30 & 14.47 & 28.84 & 27.90 & 27.74 & 26.84 & 25.22 \\
SS3D \cite{liu2022ss3d} & Sparse (2.2\%) & 49.33 & 48.40 & 41.93 & 41.03 & 35.74 & 30.98 & 29.57 & 25.88 & 40.36 & 39.19 & 38.24 & 37.54 & 38.18 \\
SS3D++ & Sparse (2.2\%) & \textbf{63.49} & \textbf{62.93} & \textbf{55.30} & \textbf{54.81} & \textbf{48.93} & \textbf{43.89} & \textbf{41.53} & \textbf{37.22} & \textbf{51.87} & \textbf{50.81} & \textbf{50.07} & \textbf{49.04} & \textbf{50.82} \\ \hline
\dag Voxel-RCNN \cite{voxelrcnn} & Sparse (2.2\%) & 47.83 & 46.32 & 41.83 & 40.51 & 25.88 & 20.73 & 21.78 & 17.44 & 50.29 & 48.48 & 48.38 & 46.65 & 38.01 \\
SS3D \cite{liu2022ss3d} & Sparse (2.2\%) & 50.74 & 49.38 & 45.11 & 44.47 & 40.56 & 36.01 & 37.94 & 34.03 & 53.27 & 51.35 & 50.88 & 48.76 & 45.20 \\
SS3D++ & Sparse (2.2\%) & \textbf{65.99} & \textbf{65.29} & \textbf{57.50} & \textbf{56.90} & \textbf{53.44} & \textbf{48.89} & \textbf{46.41} & \textbf{42.37} & \textbf{60.13} & \textbf{58.93} & \textbf{57.40} & \textbf{56.26} & \textbf{55.79} \\ \hline
\end{tabular}}
\label{tab:waymo_result_ss3d}
\end{table*}

\section{Computational Overhead Comparison}

{
To analysis the trade off between SS3D and SS3D++, we compare the computational overhead of SS3D and SS3D++ on both the Waymo and KITTI datasets. All experiments are conducted on the Tesla A100 (40GB) GPU, paired with an Intel(R) Xeon(R) Gold 5318Y @ 2.10GHz CPU. The software environment includes PyTorch version 1.10.0 and CUDA version 11.3. As shown in Tab. \ref{tab:memory_com}, the computational overhead comparison reveals that SS3D++ incurs a slight increase in training time and computational cost compared to SS3D. Notably, for the Waymo dataset, SS3D++ demonstrates a 23.4\% improvement in average performance, increasing from 45.20 to 55.79. However, this comes with a 16.6\% increase in training time and a 9.1\% increase in memory usage. Therefore, the substantial performance gains achieved by SS3D++ make the additional overhead well worth the improvement in results, indicating that this is an essential improvement.}

\begin{table*}[h]
\centering
\caption{Comparison of computational overhead between SS3D and SS3D++.
}
\resizebox{0.8\textwidth}{!}{
\begin{tabular}{c||ccc|ccc}
\hline
\multirow{2}{*}{Metric} & \multicolumn{3}{c|}{Waymo} & \multicolumn{3}{c}{KITTI} \\
 & SS3D & SS3D++ & Change (\%) & SS3D & SS3D++ & Change (\%) \\ \hline
Training Time (hours) & 8.4h & 9.8h & $\Delta$16.6\% & 15.9 & 16.8h & $\Delta$5.6\% \\
Memory Usage (GB) & 60.3GB & 65.8GB & $\Delta$9.1\% & 30.7GB & 32.3GB & $\Delta$5.2\% \\
Training Resources & 4 $\times$ A100 & 4 $\times$ A100 & $\Delta$0\% & 2 $\times$ A100 & 2 $\times$ A100 & $\Delta$0\% \\
Backbone Detector & Voxel-RCNN & Voxel-RCNN & $\Delta$0\% & PV-RCNN & PV-RCNN & $\Delta$0\% \\
Performance Avg. & 45.20 & 55.79 & $\Delta$23.4\% & 80.87 & 82.89 & $\Delta$2.5\% \\ \hline
\end{tabular}}
\label{tab:memory_com}
\end{table*}

\section{Annotation Analysis for the proposed sparsely-supervised method} \label{annotation}
\begin{figure}[h]
    \centering
    \includegraphics[width=0.99\linewidth]{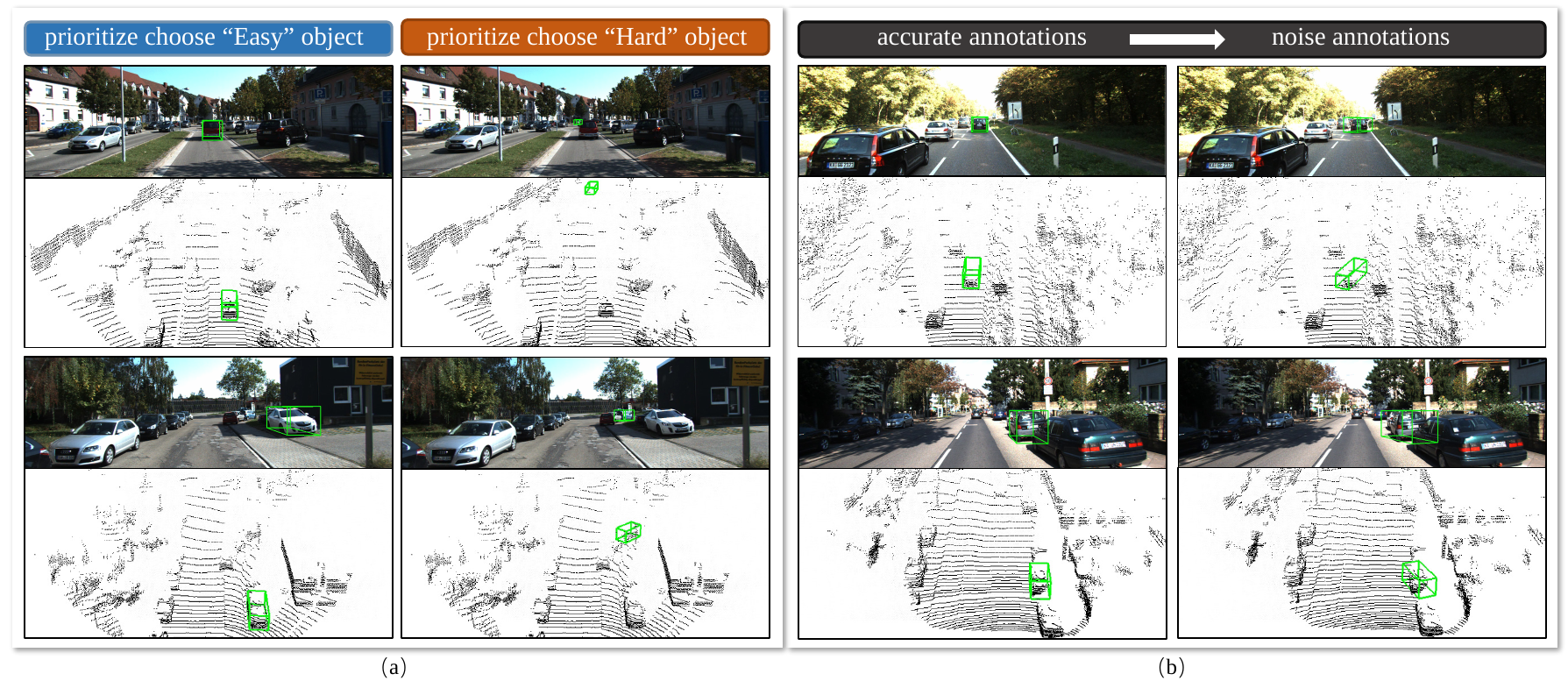}
    \caption{{(a) Illustration of the annotation process for objects with varying levels of difficulty: on the left, priority is given to annotating ``Easy'' objects, while on the right, priority is given to annotating ``Hard'' objects. Besides, with the ``Random'' setting, we can validate the impact of which box to annotate. (b) The process of introducing noise annotations by perturbing accurate annotations.}}    	  	
    \label{fig:noise_one_box} 
\end{figure}

{In our sparsely-annotated framework, we just annotate one 3D object per scene. A natural question arises: which object to annotate, and how does the accuracy of the annotations impact the results?}

{On the one hand, most annotators may be more inclined to choose the easiest object to annotate in practice, e.g., no occluded objects or high-density objects. To validate the influence of which box to annotate, we design the following three different annotated strategies: (1) randomly choose one object from the scene to annotate as default; (2) prioritize choosing one `easy' (close and no truncated) object from the scene; (3) prioritize choosing one `hard' (remote and truncated) object from the scene. Different annotated strategies are shown in Fig. \ref{fig:noise_one_box} (a).
Besides, we repeat the experiment 3 times and average the results to eliminate the influence of experimental randomness.
From Tab. \ref{tab:annotation_randomness}, we can observe that the hardness of annotated objects has a significant impact on the performance of the fully supervised detector, in which annotated strategy (3) severely deteriorates the detectors. 
However, by employing the proposed SS3D++ method, regardless of different annotated strategies, such annotation bias can be almost eliminated. Especially, the SS3D++ can achieve on-par performance compared to detectors trained with fully-supervised data for all cases. This shows that our method could perform robustly with different annotated strategies, which highlights the capability of our method to effectively address such low-quality annotations and further substantiates its potential practical value.}

\begin{table}[h]
\centering
\caption{Annotation analysis of three different sparsely-annotated forms.
``(1) (2) (3)'' denotes different sparse data settings: (1) randomly choose one object from the scene to annotate as default; (2) prioritize choosing one ``easy'' object from the scene; (3) prioritize choosing one ``hard'' object from the scene.
}
\resizebox{0.8\columnwidth}{!}{
\begin{tabular}{c|c||ccc|ccc|c}
\hline
\multirow{2}{*}{Method} & \multirow{2}{*}{Data} & \multicolumn{3}{c|}{Car - 3D Detection} & \multicolumn{3}{c|}{Car - BEV Detection} & \multirow{2}{*}{{Avg.}} \\
 &  & Easy & Mod & Hard & Easy & Mod & Hard &  \\ \hline
Voxel-RCNN & Full & 92.38 & 85.29 & 82.86 & 95.52 & 91.25 & 88.99 & {89.38} \\ \hline
Voxel-RCNN & (1) Sparse & 72.97 & 63.05 & 60.23 & 81.64 & 74.91 & 71.93 & {70.78} \\
Ours (SS3D++) & (1) Sparse & 93.50 & 84.36 & 82.64 & 96.63 & 91.31 & 89.02 & {89.57} \\
{\emph{Improvements}} & {-} & {+20.53} & {+21.31} & {+22.41} & {+14.99} & {+16.40} & {+17.09} & {+18.78}
 \\ \hline
Voxel-RCNN & (2) Sparse & 88.86 & 63.04 & 47.90 & 93.61 & 70.82 & 52.77 & {69.50} \\
Ours (SS3D++) & (2) Sparse & 92.89 & 83.46 & 79.81 & 96.40 & 91.63 & 86.73 & {88.48} \\
{\emph{Improvements}} & {-} & {+4.03} & {+20.42} & {+31.91} & {+2.79} & {+20.81} & {+33.96} & {+18.98}
 \\ \hline
Voxel-RCNN & (3) Sparse & 29.15 & 42.92 & 49.62 & 35.81 & 53.83 & 60.90 & {45.37} \\
Ours (SS3D++) & (3) Sparse & 94.24 & 84.11 & 82.44 & 96.65 & 91.38 & 88.99 & {89.63} \\
{\emph{Improvements}} & {-} & {+65.09} & {+41.19} & {+32.82} & {+60.84} & {+37.55} & {+28.09} & {+44.26}
 \\ \hline
\end{tabular}}
\label{tab:annotation_randomness}
\end{table}

{On the other hand, inaccurate annotations introduce noise to the annotations, highlighting the inherent challenges of the machine learning task. This label noise can significantly impede model training, limiting the model's ability to learn effectively. To assess the impact of annotation accuracy, as shown in Fig. \ref{fig:noise_one_box} (b), we randomly perturb some of the annotated boxes, ensuring that the IoU of the perturbed boxes with the original ground truth boxes fall within the range of $[0.45, 0.55]$. 
Additionally, we set two perturbation ratios: 20\%, and 50\%, indicating the proportion of boxes that are perturbed. 
The experimental results, as presented in Tab. \ref{tab:annotation_noise}, indicate that varying ratios of perturbation adversely affect both the fully supervised method and our approach. As the perturbation ratio increases, the detrimental effects become increasingly pronounced. However, in comparison to the fully supervised method, our approach exhibits greater robustness, consistently outperforming it across different ratios of perturbation.}

\begin{table}[h]
\centering
\caption{Annotation analysis of different perturbation ratios.}
\resizebox{0.8\columnwidth}{!}{
\begin{tabular}{c|c|c||ccc|ccc|c}
\hline
\multirow{2}{*}{Method} & \multirow{2}{*}{Data} & Perturbation & \multicolumn{3}{c|}{Car - 3D Detection} & \multicolumn{3}{c|}{Car - BEV Detection} & \multirow{2}{*}{Avg.} \\
 &  & Ratio & Easy & Mod & Hard & Easy & Mod & Hard &  \\ \hline
Voxel-RCNN & Full & 0\% & 92.38 & 85.29 & 82.86 & 95.52 & 91.25 & 88.99 & 89.38 \\
Voxel-RCNN & Sparse & 0\% & 75.55 & 64.67 & 62.43 & 83.34 & 76.75 & 73.50 & 72.71 \\
Ours (SS3D++) & Sparse & 0\% & 93.59 & 83.83 & 82.79 & 96.71 & 91.70 & 89.26 & 89.65 \\
\emph{Improvements} & - & - & +1.21 & -1.46 & -0.07 & +1.19 & +0.45 & +0.27 & +0.27 \\ \hline
Voxel-RCNN & Full & 20\% & 92.07 & 82.80 & 80.11 & 95.46 & 90.79 & 87.60 & 88.14 \\
Voxel-RCNN & Sparse & 20\% & 74.98 & 65.65 & 61.83 & 82.98 & 75.52 & 72.25 & 72.20 \\
Ours (SS3D++) & Sparse & 20\% & 92.82 & 82.86 & 79.99 & 96.33 & 91.00 & 88.44 & 88.57 \\
\emph{Improvements} & - & - & +0.75 & +0.06 & -0.12 & +0.87 & +0.21 & +0.84 & +0.44 \\ \hline
Voxel-RCNN & Full & 50\% & 92.17 & 82.39 & 79.83 & 95.64 & 88.56 & 86.98 & 87.60 \\
Voxel-RCNN & Sparse & 50\% & 73.17 & 61.79 & 58.33 & 81.17 & 74.53 & 72.06 & 70.18 \\
Ours (SS3D++) & Sparse & 50\% & 92.85 & 81.89 & 79.03 & 96.12 & 89.99 & 87.71 & 87.93 \\
\emph{Improvements} & - & - & +0.68 & -0.50 & -0.80 & +0.48 & +1.43 & +0.73 & +0.34 \\ \hline
\end{tabular}}
\label{tab:annotation_noise}
\end{table}

{It can be observed that training a detector with the original fully-supervised network under different data annotation conditions yields varying results. For instance, training the network with prioritize keeping easy objects results in higher performance for the "Easy" difficulty level but lower performance for the "Hard" difficulty level. Conversely, training the network with prioritize keeping difficult objects leads to lower performance for the "Easy" difficulty level but higher performance for the "Hard" difficulty level. However, after embedding the network into the proposed framework, regardless of the annotation condition, SS3D++ still achieves on-par performance comparing to fully-supervised method under three difficulty, indicating practical value.}

{To sum up, the two experiments mentioned above validate the robustness of our method in terms of annotation accuracy and annotation randomness, further highlighting the immense potential of our method in practical applications.}

\section{Additional Analysis of Visualization Comparison} \label{app_sec1}

\subsection{Comparison on the KITTI Dataset}
The detection results of PointRCNN~\cite{pointrcnn} and PartA$^2$~\cite{parta2} are shown above the dashed line in Fig. \ref{fig:app_kitti_vis}.
As can be seen, the proposed SS3D++ method can obviously improve the detection performance and overcome the various error cases with different detector, which further demonstrates the superiority of the detector-agnostic property.

\subsection{Comparison on the Waymo Open Dataset}
\begin{figure*}[htp]
    \centering
    \includegraphics[width=0.85\linewidth]{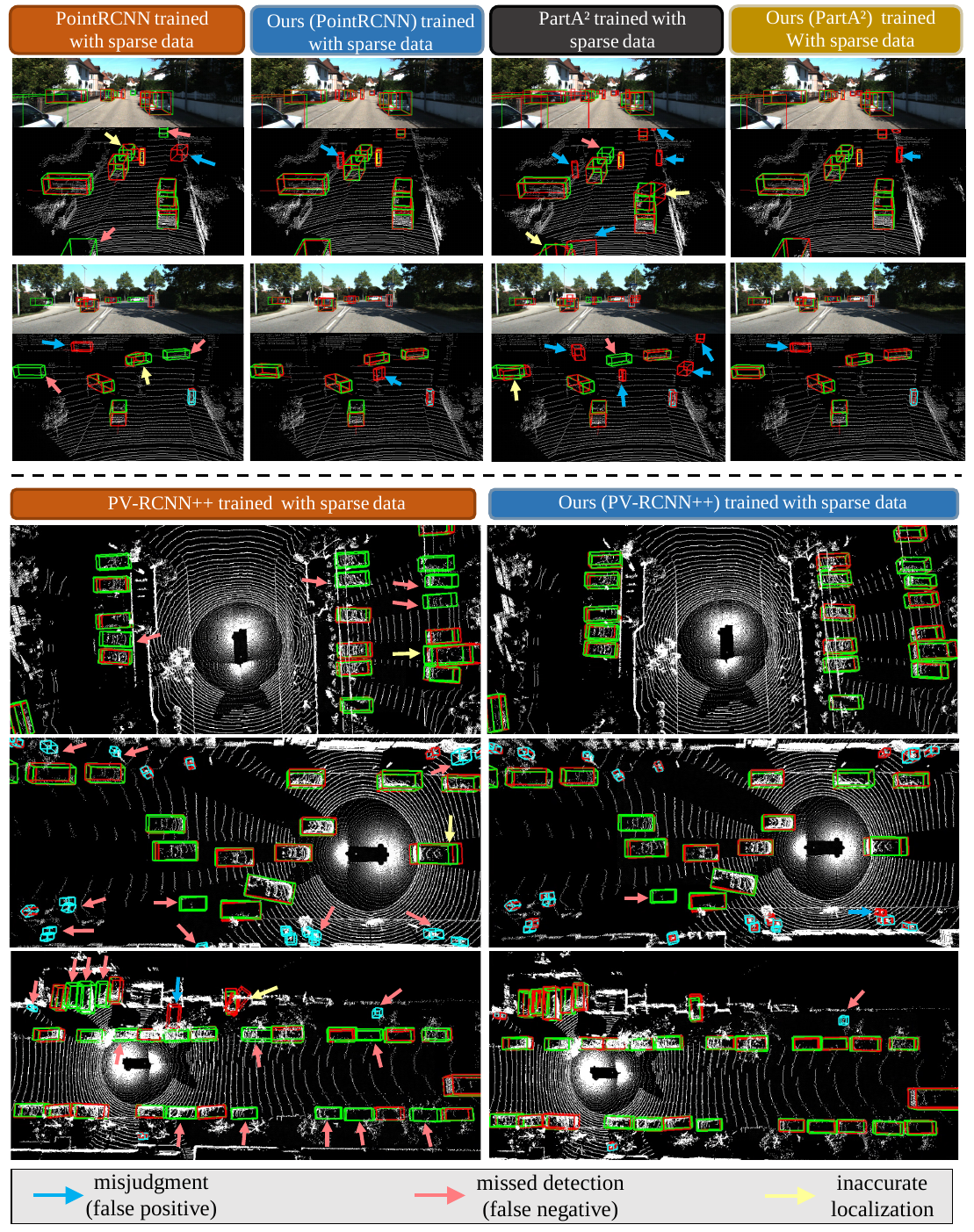}
    \vspace{-2mm}
    \caption{The qualitative comparison results on the KITTI dataset of the PointRCNN~\cite{pointrcnn}/PartA$^2$~\cite{parta2} and our SS3D++ method trained with sparsely-annotated data are shown above the dashed line, while the qualitative comparison results on the Waymo dataset of the PV-RCNN++~\cite{pvrcnn++} and our SS3D++ method trained with sparsely-annotated data are shown below the dashed line.
    The ground truth 3D bounding boxes of cars, cyclists, and pedestrians are drawn in green, yellow, and cyan, respectively. 
    For the KITTI, we set the predicted bounding boxes in red and project boxes in point cloud back onto the color images for visualization.}
    \label{fig:app_kitti_vis} 
\end{figure*}

\begin{figure*}[htp]
    \centering
    \includegraphics[width=0.85\linewidth]{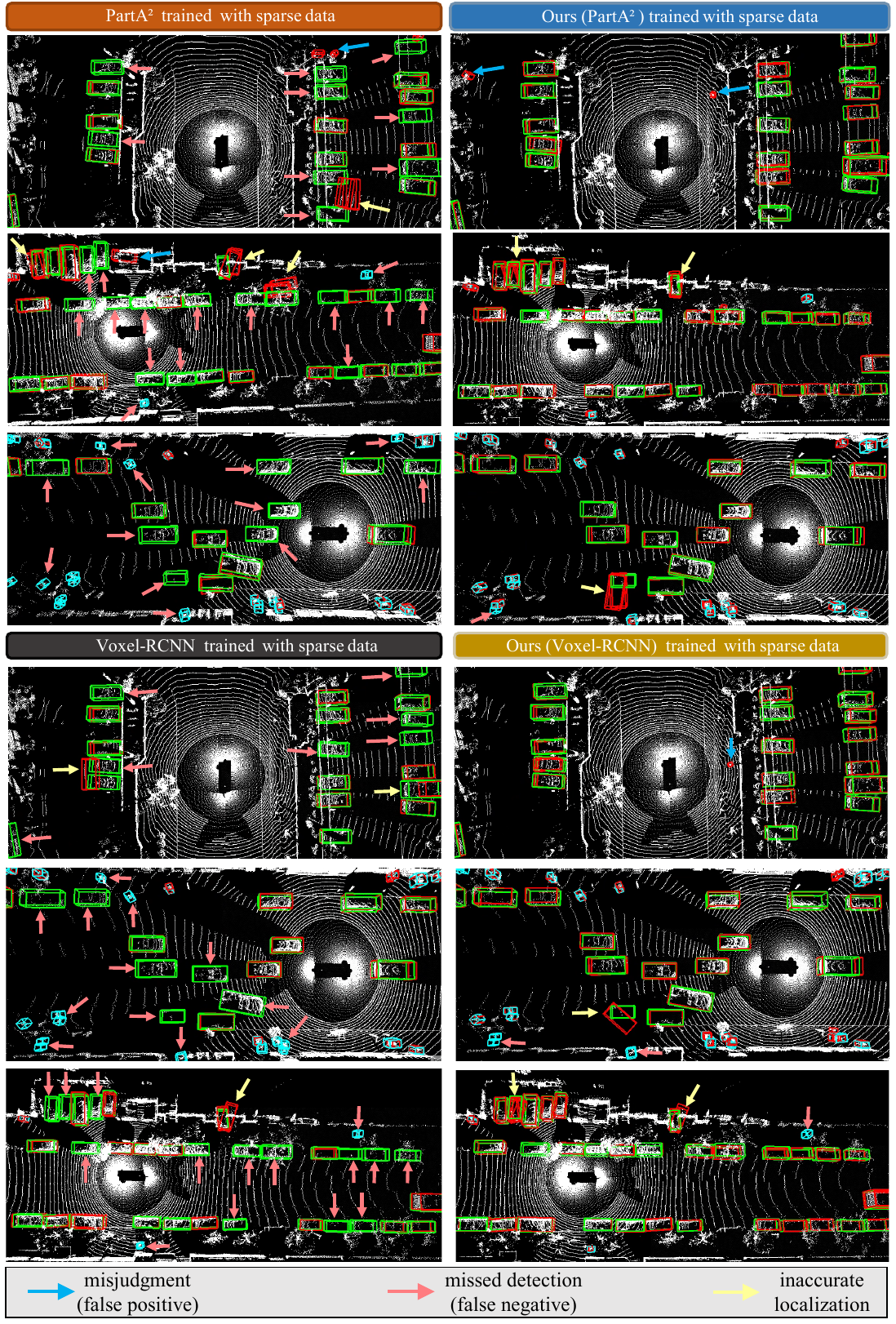}
    \vspace{-2mm}
    \caption{Qualitative comparison results of the original PartA$^2$~\cite{parta2}/Voxel-RCNN~\cite{voxelrcnn} and our SS3D++ method trained with sparsely-annotated data on the Waymo Open dataset. 
    The ground truth 3D bounding boxes of car and pedestrian are drawn in green and cyan, respectively. 
    And we set the predicted bounding boxes in red.}
    \label{fig:app_waymo_vis} 
\end{figure*}

To further verify the efficiency of our SS3D++ on the large-scale 3D autonomous driving dataset, we qualitatively show the visualization achieved by different approaches below the dashed line in Fig. \ref{fig:app_kitti_vis} on the validation set of the Waymo dataset. 
Clearly, the SS3D++ method can effectively detect more foreground objects and prevent the missed detection, especially for the objects with small sizes (e.g., pedestrian), as well as objects far away from the LiDAR sensors. Moreover, from the analysis of the two different scenarios in the KITTI and Waymo datasets, it can be observed that directly training the original detector on sparsely annotated dataset leads to significant variations in detection errors. For example, the most common error in the KITTI dataset is false positive detection (i.e., detecting the 
background as foreground shown in
Fig. \ref{fig:app_kitti_vis}), while in the Waymo dataset, the most common error is missed detection. However, our SS3D++ method greatly reduces the occurrence of these errors, further demonstrating the generalization capability of our approach. The visualization of other detectors on the Waymo can be find in Fig. \ref{fig:app_waymo_vis}.

\section{Additional Analysis of Pseudo Instances} \label{app_sec2}
\begin{figure*}[htp]
    \centering
    \includegraphics[width=\linewidth]{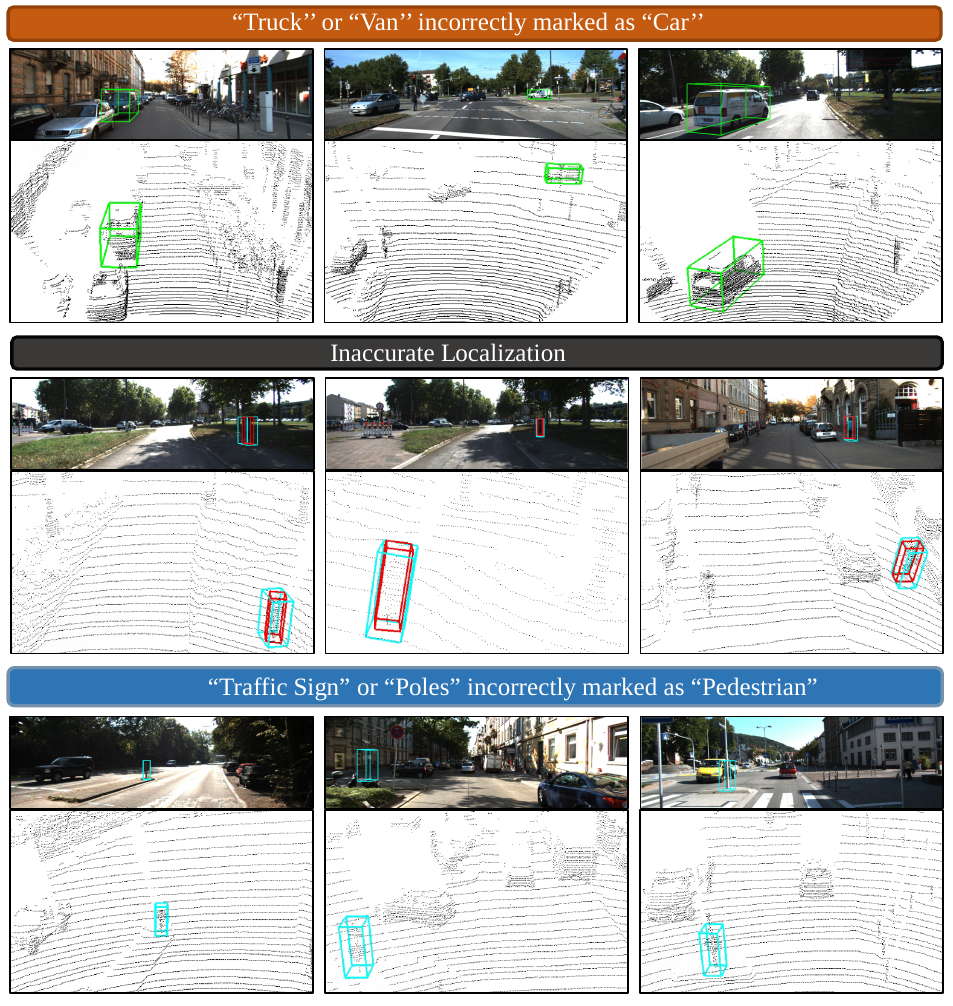}
    \caption{Visualization of failure cases of pseudo instances during the SS3D++ training process on 2\% labeled data. Pseudo instance of car and pedestrian are drawn in green and red, respectively. The ground truth of pedestrian is drawn in cyan.}
    \label{fig:app_kitti_error_pseudo} 
\end{figure*}

\begin{figure*}[htp]
    \centering
    \includegraphics[width=\linewidth]{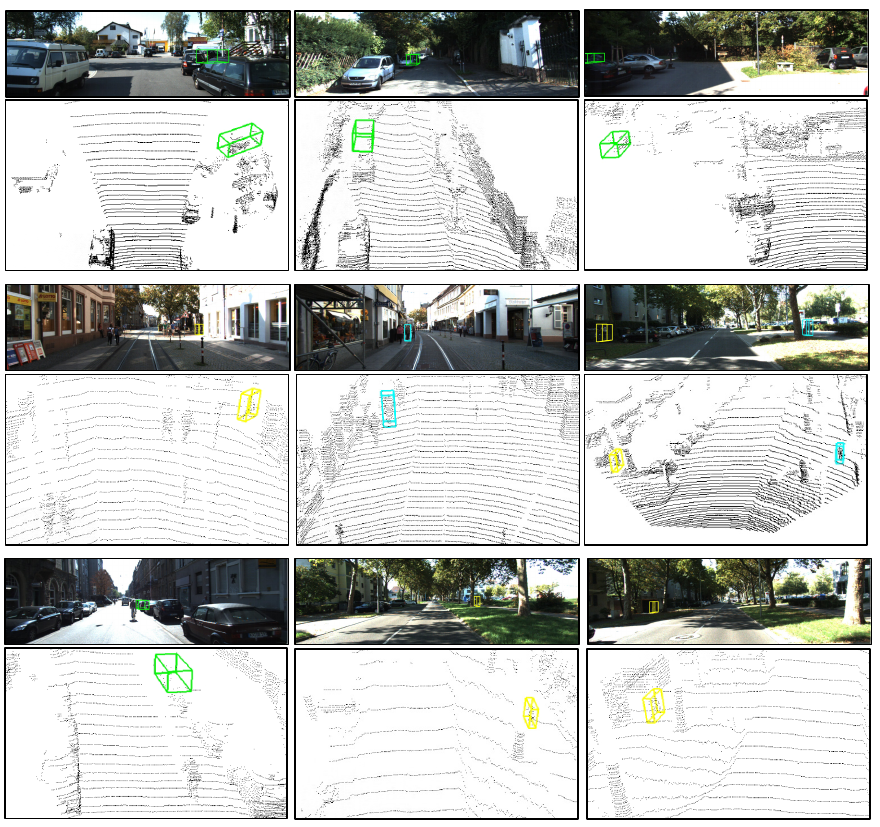}
    \caption{Visualization of interesting cases of pseudo instances during the SS3D++ training process on 2\% labeled data. The pseudo instance 3D bounding boxes of car, cyclist, and pedestrian are drawn in green, yellow, and cyan, respectively.}
    \label{fig:app_kitti_interest_pseudo} 
\end{figure*}

To intuitively explore the influence of pseudo instances, we provide qualitative results of failure pseudo instance cases during the training process on 2\% KITTI dataset, and the results are illustrated in Fig. \ref{fig:app_kitti_error_pseudo}.
For ease of viewing, we only show the key instances that need to be analyzed.
As we can seen, the pseudo instances of car categories is usually accurate, and common errors occur in categories that are relatively similar, such as trucks and vans (see the top box in Fig. \ref{fig:app_kitti_error_pseudo}).
Besides, due to the small spatial size of pedestrians, it is difficult to extract corresponding features, so the common error case is low IoU caused by inaccurate localization (the middle box in Fig. \ref{fig:app_kitti_error_pseudo}). A similar situation occurs for cyclists. Additionally, as the bottom box shown in Fig. \ref{fig:app_kitti_error_pseudo}, traffic signs and poles with similar point cloud features are mistaken for pedestrians.
In our future work, addressing these errors will be the focus of our research.
More interestingly, our pseudo instances can also identify some instances that were manually omitted in the original dataset, as shown in Fig. \ref{fig:app_kitti_interest_pseudo}. This situation of missed annotations typically occurs due to overcrowding of objects (such as occlusion or truncation) or insufficient lighting conditions, which may lead to the objects being overlooked by the annotator. Hence, this discovery highlights the tremendous potential of our method as an annotation tool to enhance the quality of annotations.

\section{Additional Analysis of Learning from Unlabeled Training Scenes in An Online Manner} \label{app_sec3}
For an autonomous driving system in practice, the 3D detection needs to  perceive the changing environment with non-stationary data distribution and incrementally improve itself. Algorithm \ref{alg:stream_input} summarizes that SS3D++ method learns from unlabeled training scenes in an online manner. In our implementation, we take Voxel-RCNN~\cite{voxelrcnn} as our base detector and initialize the SS3D++ method with 200 sparsely annotated sequential point cloud scenes.

\renewcommand{\algorithmicrequire}{ \textbf{Input:}} 
\renewcommand{\algorithmicensure}{ \textbf{Output:}} 
\begin{algorithm}[h]
    \caption{Algorithm of Learning from Streaming Unlabeled Scenes.}
    \label{alg:stream_input}
    \begin{algorithmic}[1] 
    \REQUIRE $\mathcal{S}_t$: stream unlabeled scenes at time $t$, $\mathcal{M}$: scenes stored in memory, $T$: the number of streaming tasks, 3D detector $F$ with weight parameter $\phi$.\\
    \STATE Initialize 3D detector $F$ using the SS3D++ method with initial sparsely annotated scenes and update weight parameter $\phi$;
    \FOR{ $t=1$ to $T$ }
        \STATE Fetch current streaming unlabeled scenes $\mathcal{S}_t$ and memory scenes $\mathcal{M}$; 
        \STATE Alternatively improve 3D detector training and confident fully-annotated scene generation using the SS3D++ method based on $\mathcal{S}_t \cup \mathcal{M}$.
        \STATE Update memory $\mathcal{M}$ by randomly selecting generated full-annotated scenes, and remove the remaining training scenes.
    \ENDFOR
\ENSURE Learned 3D detector $F$
\end{algorithmic}
\end{algorithm}

\section{Additional Analysis of Utilizing the Temporal Information}

\begin{figure*}[h]
    \centering
    \includegraphics[width=0.6\linewidth]{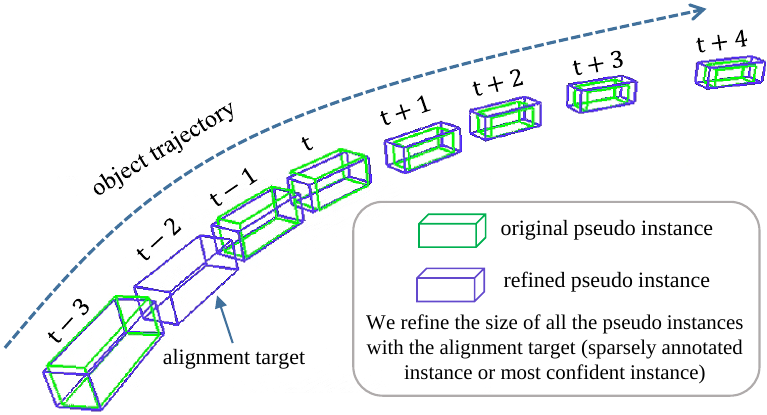}
    \caption{{Illustration of pseudo instance refinement using temporal information.}}    	
    \label{fig:refined_box_time} 
\end{figure*}

{In autonomous driving data, temporal information is abundant, and the methods for leveraging this temporal information have been progressively studied.
It can be observed that for rigid objects, their geometric shapes remain largely unchanged across consecutive time periods, regardless of their motion state. Furthermore, the motion state of an object typically demonstrates regularity and strong consistency with adjacent moments. Inspired by \cite{ma2023detzero,qi2021offboard}, our method can easily integrate temporal information for further pseudo instance refinement.
Specifically, after obtaining pseudo instances through confident missing-annotated instances mining module, we can effectively enhance their accuracy using temporal information. As suggested by the review, we attempt to leverages this information to refine pseudo instances by ensuring consistency across the sequence of point clouds, and the algorithm is shown in Algorithm \ref{alg:refine}. To facilitate understanding, we have also illustrated the process in Fig. \ref{fig:refined_box_time}. Next, we will provide a detailed description of the process for refining pseudo instances using temporal information.}

\renewcommand{\algorithmicrequire}{ \textbf{Input:}} 
\renewcommand{\algorithmicensure}{ \textbf{Output:}} 
\begin{algorithm}[h]
	\caption{{Algorithm of pseudo instance refinement using temporal information.}}
	\label{alg:refine}
	\begin{algorithmic}[1] 
		\REQUIRE 3D instances from the instance bank ${\mathcal{B}}^{(t)}$, sequence of $N$ point cloud scenes $\{P_i\}, i=1,2,\dots,N$;\\
		\STATE Fetch 3D bounding boxes $b_{ij}$, class $y_{ij}$, confidence scores $s_{ij}$ ($j$-th instance at frame $i$) from sequence $\{P_i\}$ based on the instance bank ${\mathcal{B}}^{(t)}$;
		\STATE Perform multi-object tracking algorithm to obtain set of tracks $\mathcal{K}$ and link 3D instances across frames;
		\FOR { $k=0$ to $\mathcal{K}$ }
		\STATE Fetch 3D bounding boxes $\{b_{ijk}\}$ and corresponding confidence score $\{s_{ijk}\}$ from object track $k$;
            \IF{initial sparsely annotated instance in $\{b_{ijk}\}$}
                \STATE Refine the bounding boxes $\{b_{ijk}\}$ geometry size to align with the shape of the sparsely annotated instance;
            \ELSE
                \STATE Find the most confident pseudo instance according to the confidence score $\{s_{ijk}\}$;
                \STATE Refine the bounding boxes $\{b_{ijk}\}$ geometry size to align with shape of the most confident pseudo instance;
            \ENDIF
		\ENDFOR
		\ENSURE Refined instance bank ${\mathcal{B}}^{(t)}$. \\ 
	\end{algorithmic}
\end{algorithm}


{First, we implement a variant of object tracking algorithm \cite{weng2019baselinetrack} to associate objects across adjacent frames in a non-parametric manner, ensuring the consistency for the same object at different time points. Following \cite{qi2021offboard}, we transform all bounding boxes to the world coordinate for tracking, there by minimizing the impact of sensor ego-motion. Our carefully designed multi-criteria sample selection process effectively reduces the occurrence of false positive bounding boxes prior to tracking. For track association, we utilize BEV (Bird's Eye View) boxes, setting the IoU threshold of the Hungarian algorithm to 0.1.}

{Then, we can progressively update the pseudo instances in the instance bank using the aforementioned tracking information. On the one hand, if a pseudo instance corresponds to the same object (\emph{i.e.,} the same track ID) as our initial sparsely annotated instance, both instances should share the same geometric size. Therefore, we directly refine the size of all pseudo instances within the track to align with the sparsely annotated instance. On the other hand, if all instances in a track are pseudo annotated, we select the most confident pseudo instance as the alignment target by comparing their confidence scores. The pseudo instance refinement algorithm is embedded after line 7 in Algorithm 1 of the main paper, i.e., after obtaining the updated instance bank. It can be seamlessly integrated with the original method.}

{As shown in Tab. \ref{tab:waymo_result_temporal}, experimental results demonstrate that incorporating temporal information significantly improves the performance of our method on the Waymo dataset. It is worth noting that we have employed a straightforward approach to utilize temporal information, and further exploration of advanced strategies could yield even better performance. This also implies that there remains a large room for further algorithm, which will be further investigated in our future research.}

\begin{table*}[h]
\centering
\caption{{Performance on the Waymo Open Dataset with 202 validation sequences for the 3D detection. When using temporal information to refine pseudo instances, our SS3D++ achieves improved performance.}
}
\resizebox{\textwidth}{!}{
\begin{tabular}{c|c||cc|cc|cc|cc|cc|cc|c}
\hline
\multirow{2}{*}{Methods} & \multirow{2}{*}{Data} & \multicolumn{2}{c|}{Veh. (LEVEL 1)} & \multicolumn{2}{c|}{Veh. (LEVEL 2)} & \multicolumn{2}{c|}{Ped. (LEVEL 1)} & \multicolumn{2}{c|}{Ped. (LEVEL 2)} & \multicolumn{2}{c|}{Cyc. (LEVEL 1)} & \multicolumn{2}{c|}{Cyc. (LEVEL 2)} & \multirow{2}{*}{Avg.} \\
 &  & mAP & mAPH & mAP & mAPH & mAP & mAPH & mAP & mAPH & mAP & mAPH & mAP & mAPH &  \\ \hline
CenterPoint \cite{centerpoint} & Sparse (2.2\%) & 32.98 & 32.33 & 28.82 & 28.24 & 20.22 & 16.91 & 17.30 & 14.47 & 28.84 & 27.90 & 27.74 & 26.84 & 25.22 \\ \hline
SS3D++ (Ours) & Sparse (2.2\%) & 63.49 & 62.93 & 55.30 & 54.81 & 48.93 & 43.89 & 41.53 & 37.22 & 51.87 & 50.81 & 50.07 & 49.04 & 50.82 \\
+ temporal information & Sparse (2.2\%) & 65.06 & 64.53 & 56.76 & 56.29 & 53.16 & 48.10 & 45.83 & 41.40 & 53.17 & 52.07 & 52.04 & 50.99 & 53.28 \\
\emph{Improvements} & - & +1.57 & +1.60 & +1.46 & +1.48 & +4.23 & +4.21 & +4.30 & +4.18 & +1.30 & +1.26 & +1.97 & +1.95 & +2.46 \\ \hline
\end{tabular}}
\label{tab:waymo_result_temporal}
\end{table*}

\end{document}